\documentclass{ecai}

\usepackage{latexsym}
\usepackage{amssymb}
\usepackage{amsmath}
\usepackage{amsthm}
\usepackage{booktabs}
\usepackage{enumitem}
\usepackage{graphicx}
\usepackage{color}

\usepackage{algorithm}
\usepackage{algpseudocode}

\usepackage{caption}

\newcommand{\BibTeX}{B\kern-.05em{\sc i\kern-.025em b}\kern-.08em\TeX}

\usepackage[dvipsnames]{xcolor}
\usepackage{multirow}
\usepackage{algorithm}
\usepackage{algpseudocode}
\usepackage{graphicx}

\newcommand{\colorM}[1]{{\color[HTML]{376795}#1}}

\algtext*{EndFor}%
\algtext*{EndIf}%

\usepackage{amsmath,amsfonts,bm}

\def\Tabref#1{Table~\ref{#1}}

\def\Figref#1{Figure~\ref{#1}}

\def\Secref#1{Section~\ref{#1}}

\def\eqref#1{equation~\ref{#1}}
\def\Eqref#1{Equation~\ref{#1}}

\def\1{\bm{1}}

\DeclareMathAlphabet{\mathsfit}{\encodingdefault}{\sfdefault}{m}{sl}
\SetMathAlphabet{\mathsfit}{bold}{\encodingdefault}{\sfdefault}{bx}{n}

\begin{document}

\begin{frontmatter}

\paperid{516}

\title{Weight Scope Alignment: A Frustratingly Easy Method for Model Merging}

\author[AB]{\fnms{Yichu}~\snm{Xu}}
\author[AB]{\fnms{Xin-Chun}~\snm{Li}}
\author[AB]{\fnms{Le}~\snm{Gan}}
\author[AB]{\fnms{De-Chuan}~\snm{Zhan}\thanks{Corresponding Author. Email: 	zhandc@nju.edu.cn}}

\address[A]{School of Artificial Intelligence, Nanjing University, China}
\address[B]{National Key Laboratory for Novel Software Technology, Nanjing University, China}

\begin{abstract}
    Merging models becomes a fundamental procedure in some applications that consider model efficiency and robustness. 
    The training randomness or Non-I.I.D. data poses a huge challenge for averaging-based model fusion. 
    Previous research efforts focus on element-wise regularization or neural permutations to enhance model averaging while overlooking weight scope variations among models, which can significantly affect merging effectiveness.
    In this paper, we reveal variations in weight scope under different training conditions, shedding light on its influence on model merging. 
    Fortunately, the parameters in each layer basically follow the Gaussian distribution, which inspires a novel and simple regularization approach named \textbf{W}eight \textbf{S}cope \textbf{A}lignment (\textbf{WSA}). 
    It contains two key components: 1) leveraging a target weight scope to guide the model training process for ensuring weight scope matching in the subsequent model merging. 2) fusing the weight scope of two or more models into a unified one for multi-stage model fusion. 
    We extend the WSA regularization to two different scenarios, including Mode Connectivity and Federated Learning. 
    Abundant experimental studies validate the effectiveness of our approach.
    \end{abstract}

\end{frontmatter}

\section{Introduction} \label{sec:intro}

As a technique for combining multiple deep models into a single model, model fusion~\citep{li2023DeepModel,jin2023DatalessKnowledge} \citep{li2023DeepModel} has gained widespread applications across various domains, including Mode Connectivity~\citep{ashmore2015MethodFinding,draxler2018EssentiallyNo,entezari2022RolePermutation} and  Federated Learning~\citep{acar2021FederatedLearning,mcmahan2017CommunicationefficientLearninga,yang2019FederatedMachine}. 
First, the model interpolation could shed light on the properties of the mode connectivity in neural networks~\citep{goodfellow2015QualitativelyCharacterizing,frankle2020RevisitingQualitatively,garipov2018LossSurfaces}. 
Then, due to data privacy protection, transmitting intermediate models across edge nodes and fusing them on the server has been the common procedure in federated learning~\citep{wang2020FederatedLearning,li2020FederatedOptimization,li2022UnderstandingFederated}. 
To be brief, model fusion matters a lot in these applications and has attracted a wide range of research interest. 
The primary goal of model fusion is to retain the capabilities of the original models while achieving improved generalization, efficiency, and robustness.

The coordinate-based parameter averaging is the most common approach for model fusion in deep neural networks~\citep{mcmahan2017CommunicationefficientLearninga,li2023DeepModel,wortsman2022ModelSoups}. 
The research on mode connectivity involves a linear or piece-wise interpolation between models~\citep{goodfellow2015QualitativelyCharacterizing,vlaar2022WhatCan}, while federated learning takes the averaging of local models from edge nodes for aggregation~\citep{mcmahan2017CommunicationefficientLearninga,dinh2020PersonalizedFederated}. 
Although parameter averaging exhibits favorable properties, it may not perform optimally in more complex training scenarios, especially when faced with various training conditions or Non-Independent and Identically Distributed (Non-I.I.D.) data. 
For example, the Non-I.I.D. data in federated learning means that the data of local nodes are naturally heterogeneous, making the model aggregation suffer from diverged update directions~\citep{hsieh2020NonIIDData,jeong2018CommunicationefficientOndevice}. 
Additionally, the property of permutation invariance that neural networks own exacerbates the challenge of model fusion because of the neuron misalignment phenomenon~\citep{entezari2022RolePermutation,yurochkin2019BayesianNonparametric,brea2019WeightspaceSymmetry,godfrey2022SymmetriesDeep}. 
Hence, solutions have been proposed from the aspect of element-wise regularization~\citep{li2020FederatedOptimization,acar2021FederatedLearning,dinh2020PersonalizedFederated} or mitigating the permutation invariance~\citep{li2022FederatedLearning,ainsworth2023GitRebasin,singh2020ModelFusion,pittorino2022DeepNetworks}. 
Few of these methods, however, have considered the impact of weight ranges across models on model fusion.

\begin{figure}[tbp]
    \centering
    \includegraphics[width=\linewidth]{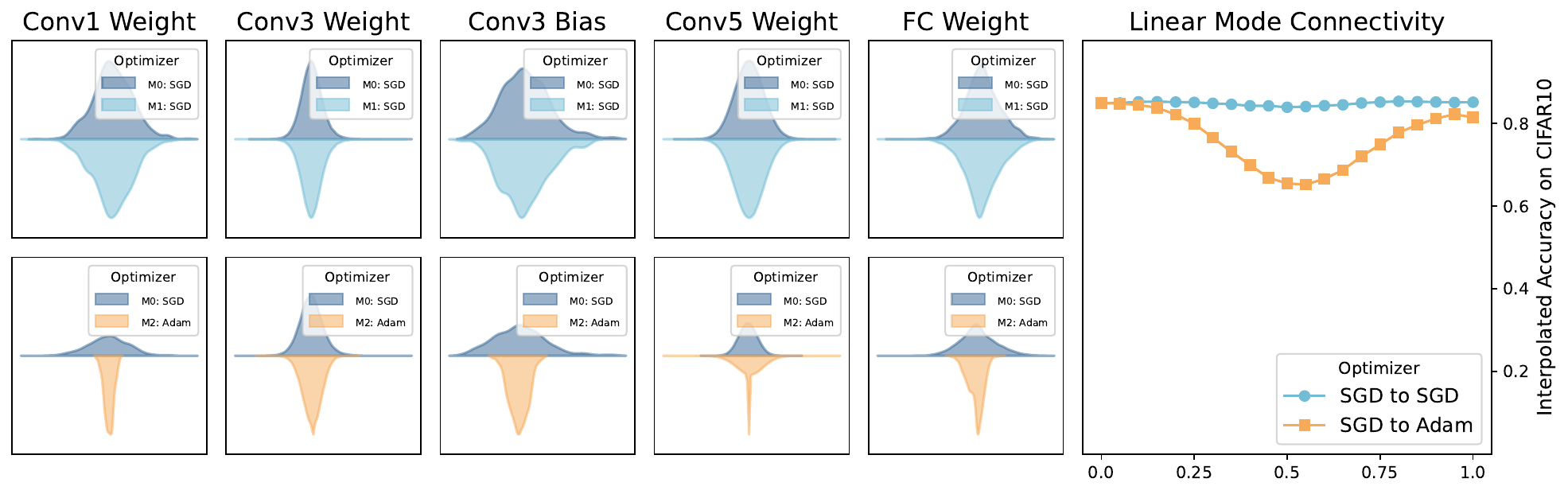}
    \caption{\textbf{Left}: Similar optimizers yield similar weight distributions. Conversely, different optimizers lead to distinct weight profiles. 
    \textbf{Right}: Models trained from different training conditions (e.g., the used optimizer) tend to produce poorer model interpolations.
    Details are in \Secref{sec:obser}.}
    \label{fig:scope-mc-demo}
\end{figure}

\begin{figure*}[tbp]
    \centering
    \includegraphics[width=0.8\linewidth]{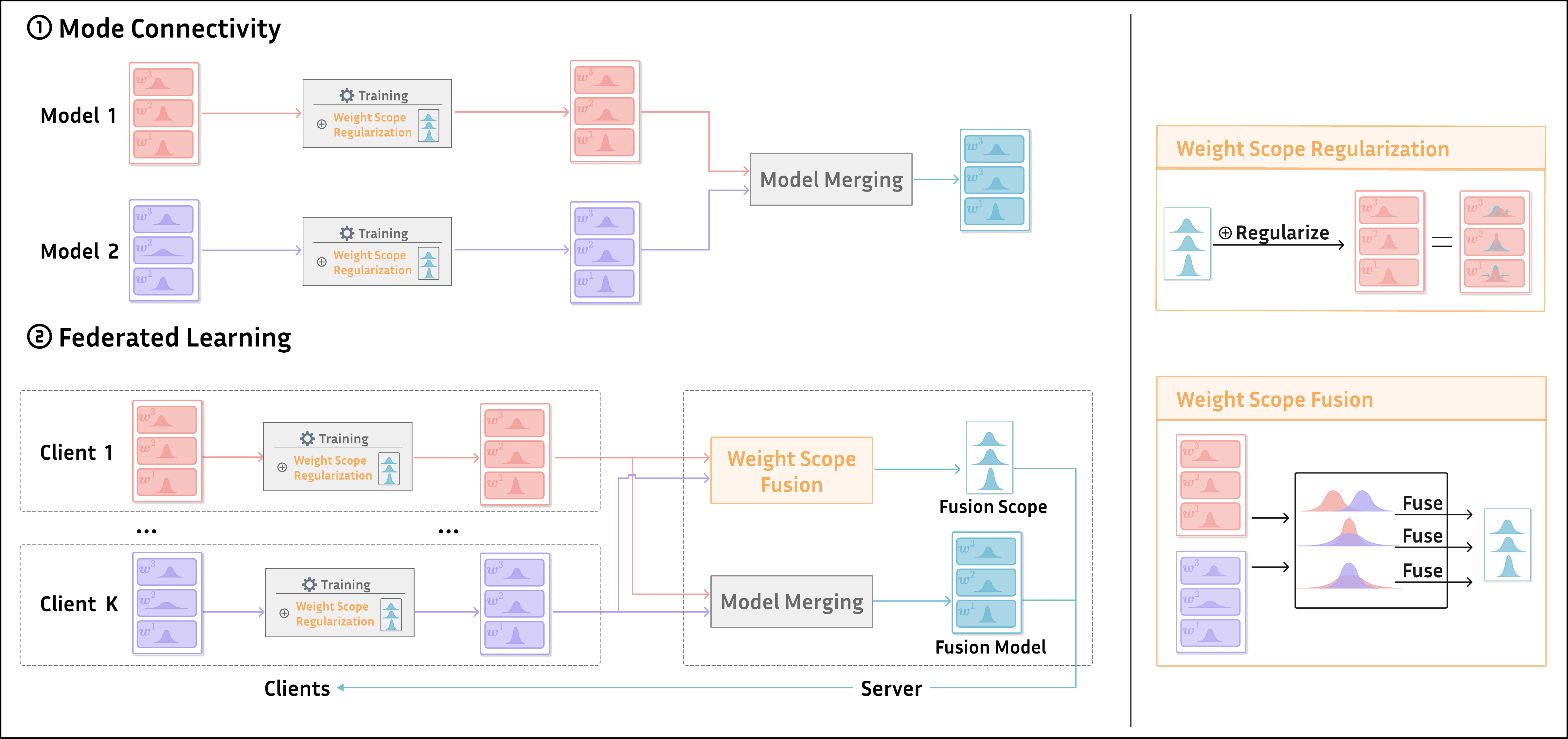}
    \caption{The Framework of Weight Scope Alignment and Its Applications. \textbf{Left}: This method can be adapted to various applications of model merging, such as Model Connectivity and Federated Learning. Mode Connectivity includes a single model fusion, while Federated Learning requires multi-stage fusion. \textbf{Right}: The method comprises two components: weight scope regularization and weight scope fusion. }
    \label{fig:fig2_framework}
\end{figure*}

In this paper, we investigate the influence of different training conditions on model weight distributions (defined as Weight Scope) and further study model merging under various weight scopes. 
We first conduct several experiments under various training hyper-parameters or data quality conditions and find that the weight scopes of the converged models differ a lot, a phenomenon we define as ``weight scope mismatch''. 
\Figref{fig:scope-mc-demo} illustrates the model weight distributions under different training conditions, revealing noticeable differences despite all distributions being approximated by Gaussian distributions. 
Specifically, the top five sub-figures are parameters from models that use the same optimizer, while the bottom ones take different optimizers. 
In the rightmost of \Figref{fig:scope-mc-demo}, the linear interpolation results that reflect the mode connectivity property are also provided. 
Clearly, the mismatched weight scope leads to a worse linear interpolation, highlighting the impact of weight range inconsistency on model fusion. 
To intuitively explain, parameters with similar distributions can be aggregated more easily, whereas those with dissimilar distributions often present challenges in model merging.

Fortunately, the parameters in each layer follow a very simple distribution, i.e., Gaussian distribution. 
The simple distribution inspires us a novel and easy way of aligning parameters. 
We leverage a target weight scope to guide the model training process, ensuring weight alignment and scope matching in subsequent model merging. 
For more complex multi-stage fusion, we calculate the mean and variance of parameter weights in the to-be-merged models, then aggregate these statistics into unified one as the target weight scope. 
The proposed method is named \textbf{W}eight \textbf{S}cope \textbf{A}lignment (\textbf{WSA}), and the above steps are named weight scope regularization and weight scope fusion, respectively. 
The whole procedure is illustrated in \Figref{fig:fig2_framework}. 
We apply WSA to scenarios of mode connectivity and federated learning for exploring the advantages of WSA compared with corresponding solutions in these areas. 
Our work aims to align the weights as closely as possible with a given distribution during training, thereby enhancing the match in weight ranges and facilitating model merging. 
Compared with other similar regularization methods, e.g., the weight decay and the proximal term, the proposed WSA seeks to balance specificity and generality, addressing the shortcomings of existing methods while optimizing for effective model fusion. 
Our contributions are as follows: {\it 1) as far as we know, our work is the first to formally study the impact of weight scope on model fusion; 2) the proposed WSA is simple yet effective, which is verified on two applications via abundant experimental studies.}

\section{Related Works} \label{sec:relate}
Model fusion is a fundamental technique in several applications. The related scenarios and solutions are introduced as follows.

\paragraph{Model Merging in Mode Connectivity.} \label{sec:relate-mc}
Visualizing loss landscape is an intuitive way to understand the mode connectivity~\cite{li2018VisualizingLoss,hao2019VisualizingUnderstanding,Li2024ExploringAE}, where the landscape is shown in a 2-dim or 3-dim space via model interpolations. The model interpolations between the initialization and the converged model could reflect the monotonic linear interpolation phenomenon~\cite{goodfellow2015QualitativelyCharacterizing,frankle2020RevisitingQualitatively,wang2023PlateauMonotonic,tao2024mli}. \cite{frankle2020LinearMode} bridges the model connectivity and lottery ticket hypothesis~\cite{frankle2019LotteryTicket}, proposing a method for model pruning. Mode connectivity is also related to the model optimization and generalization~\cite{keskar2017LargebatchTraining,dinh2017SharpMinima}. 

The work~\cite{goodfellow2015QualitativelyCharacterizing} also points out that two independent minima suffer a barrier in their linear interpolation, which attracts several solutions to decrease the barrier. \cite{draxler2018EssentiallyNo,garipov2018LossSurfaces} make a notable discovery that the independent minima are possible to be connected via a simple piece-wise or quadratic curve. \cite {neyshabur2020WhatBeing,wortsman2022ModelSoups} find that the minima fine-tuned from the same pre-trained model could mitigate the barrier in linear interpolation. \cite{entezari2022RolePermutation} guesses that the independent minima are located in the same basin with the consideration of permutation invariance, conjecturing that the minima matched via the simulated annealing algorithm encounter no barrier. \cite{singh2020ModelFusion} propose the weight-based and activation-based matching method via the optimal transport. \cite{tatro2020OptimizingMode} decreases the barrier via both the permutation alignment and the quadratic curve. \cite{ainsworth2023GitRebasin} employs three distinct neuron-matching methods to corroborate the low-barrier hypothesis. 
Although the permutation invariance is considered in these works, the mismatch of weight scope in neural networks is also fundamental to model fusion. Our proposed method could further decrease the barrier on the basis of these works.

\paragraph{Model Merging in Federated Learning.} \label{sec:relate-fl}
Federated learning (FL) aims to break the limitation of data privacy, utilizing a server that collaborates with client devices to train a model~\cite{yang2019FederatedMachine}. 
As the most standard algorithm in FL, FedAvg~\cite{mcmahan2017CommunicationefficientLearninga} takes a simple coordinate-based parameter averaging on the server to accomplish the model fusion process. 
A huge challenge that FL faces is the Non-I.I.D. data~\cite{hsieh2020NonIIDData,karimireddy2020SCAFFOLDStochastic}, where the inherent data heterogeneity leads to weight divergence during local training~\cite{jeong2018CommunicationefficientOndevice}. 
Applying coordinate-wise regularization on local models is a popular solution to solve the Non-I.I.D. challenge in FL. 
\cite{li2023FedNARFederated} claims that weight decay~\cite{krogh1991SimpleWeight} can lead to divergent optimization objectives among Non-I.I.D. clients in FL. 
\cite{zhang2015DeepLearning,li2020FederatedOptimization} introduce a proximal term in local optimization. 
This term helps align local optimization more closely with given model parameters, facilitating model aggregation. Additionally, \cite{karimireddy2020SCAFFOLDStochastic} implements constraints on the local gradient directions of each client, nudging them closer to a global direction. 
However, the coordinate-wise regularization tends to be overly specific, which may induce perturbations to model training. 

Additionally, the permutation invariance could also make the local models misaligned in FL, which is harder for aggregation. 
\cite{yurochkin2019BayesianNonparametric, wang2020FederatedLearning} consider the permutation invariance of neural networks, rearranging the order of neurons can result in multiple models with the same functionality as the original network. 
\cite{yu2021Fed2Featurealigned} introduces a group alignment method, while \cite{li2022FederatedLearning} designs the position-aware neurons to align parameters. 
These works mainly focus on the permutation invariance in FL, but they have not analyzed the influence of different model weight ranges on model fusion.

\section{Proposed Methods} \label{sec:method}
\subsection{Preliminaries} \label{sec:method-prelim}
We introduce a collection of models, denoted as $\mathcal{K}$, with the size of the collection $ |\mathcal{K}| \geq 1 $. 
Each model within this collection is constructed based on the same underlying model architecture. 
The distinctiveness among the models in $\mathcal{K}$ arises from the variations in their weight layers, which are derived from training under various hyper-parameters or different data. 

To encapsulate the statistical characteristics of the weight layers across different models, we denote the weight matrix of the $\ell^{th}$ layer in the $k^{th}$ model as $ \boldsymbol{w}_{k}^{\ell} $. 
Our assumption is that the elements within this matrix follow a Gaussian distribution which is characterized by its mean $ \mu_{k}^{\ell} $ and standard deviation $ \sigma_{k}^{\ell} $.
The assumption is rational and we will empirically present the weight distributions of the converged model in \Secref{sec:obser}.
Formally, the distribution of the weight matrix $ \boldsymbol{w}_{k}^{\ell} $ is as follows: 
\begin{equation}
p(\boldsymbol{w}_{k}^{\ell}) = \mathcal{N}(\mu_{k}^{\ell}, (\sigma_{k}^{\ell})^{2}),
\end{equation}
\begin{small}
\begin{equation}
\mu_{k}^{\ell} = \frac{1}{|\boldsymbol{w}_{k}^{\ell}|}\sum_{w \in \boldsymbol{w}_{k}^{\ell}} w, \,\,\,\, \sigma_{k}^{\ell} = \sqrt{\frac{1}{|\boldsymbol{w}_{k}^{\ell}|}\sum_{w \in \boldsymbol{w}_{k}^{\ell}} (w - \mu_{k}^{\ell})^2}. \label{eq:scope-sigma}
\end{equation}
\end{small}

In \Eqref{eq:scope-sigma}, we apply Maximum Likelihood Estimation to derive the standard deviation $\sigma_{k}^{\ell}$ and mean $\mu_{k}^{\ell}$ of the weight $\boldsymbol{w}_{k}^{\ell}$.

\subsection{Weight Scope Alignment} \label{sec:wsa}
Firstly, we introduce Weight Scope Regularization. Then, for more complex multi-stage fusion, we propose Weight Scope Fusion for better target alignment.

\paragraph{Weight Scope Regularization.} \label{sec:wsa-wsr} 
Given a weight distribution $\mathcal{N}(\mu, \sigma^{2})$ and a target weight distribution $\mathcal{N}(\tilde{\mu}, \tilde{\sigma}^{2})$, our method endeavors to ensure consistency between them.
This consistency is crucial for guaranteeing scope-matched distribution among the new models in the subsequent model fusion.
To achieve this, we employ a divergence measure known as the Kullback-Leibler (KL) divergence, specifically focusing on calculating the divergence between the two univariate Gaussian distributions, which is given by:
\begin{small}
\begin{equation}
D_{\mathrm{KL}} = \log\left(\frac{\tilde{\sigma}}{\sigma}\right) + \frac{\sigma^2 + (\mu - \tilde{\mu})^2}{2\tilde{\sigma}^2} - \frac{1}{2}, \label{eq:distribution_kl}
\end{equation}
\end{small}

where $\tilde{\mu}, \tilde{\sigma}$ are hyperparameters. By minimizing the KL divergence between the training models’ weight distribution and their goal, it ensures that the weight distributions are closely aligned, facilitating more effective and harmonious model fusion.

\paragraph{Weight Scope Fusion.}\label{sec:wsa-wsf}
In some complex scenarios with large amounts of models and multiply stages of model merging, a given pre-defined weight distribution is not adaptable.
Therefore, we propose a method named Weight Scope Fusion to enhance the applicability of Weight Scope Regularization.
Focusing on the weights of the $\ell^{th}$ layer, we assume that each weight $\boldsymbol{w}_{k}^{\ell}$, $k \in \mathcal{K}$, follows its own Gaussian distribution, and they are independent of each other.
We can get the fused Gaussian distribution $\mathcal{N}(\tilde{\mu}^{\ell},(\tilde{\sigma}^{\ell})^2)$ by:

\begin{small}
\begin{equation}
    \tilde{\mu}^{\ell} = \frac{1}{|\mathcal{K}|}\sum_{k \in \mathcal{K}} \mu_{k}^{\ell}, \,\,\,\, \tilde{\sigma}^{\ell} = \frac{1}{\sqrt{|\mathcal{K}|}} \sqrt{\sum\limits_{k \in \mathcal{K}} (\sigma_{k}^{\ell})^2}. \label{eq:distribution_fusion}
\end{equation}
\end{small}

\subsection{Analysis and Comparisons with Other Methods} \label{sec:ana}
We next provide some analysis of the proposed simple method, especially the comparisons with existing works.

\paragraph{Comparison with Weight Decay.} \label{sec:ana-decay}
Weight decay~\cite{krogh1991SimpleWeight} stands as a cornerstone in model regularization, advocating for weight constraints that push the weights towards zero and promote uniformity. 
Considering a weight vector, denoted as $\boldsymbol{w}\in \mathbb{R}^n$, with elements $[\boldsymbol{w}_1, \boldsymbol{w}_2, \dots, \boldsymbol{w}_n]$. 
For ease of analysis, we define the mean, standard deviation, and L2 norm of this vector as follows: $\mu = \frac{1}{n} \sum_{i=1}^{n} \boldsymbol{w}_i$, $\sigma = \sqrt{\frac{1}{n} \sum_{i=1}^{n} (\boldsymbol{w}_i - \mu)^2}$, $||\boldsymbol{w}||_2 = \sqrt{\sum_{i=1}^{n} \boldsymbol{w}_i^2}$. 
The weight decay term is formulated as $\frac{\lambda}{2} ||\boldsymbol{w}||_2^2$. 
We can express weight decay in terms of $\mu$ and $\sigma$ as follows:

\begin{small}
\begin{equation}
\frac{\lambda}{2} ||\boldsymbol{w}||_2^2 = \frac{\lambda n}{2} (\sigma^2 + \mu^2).
\end{equation}
\end{small}

The weight decay term brings the mean and variance of model weights close to zero, and several research \citep{acar2021FederatedLearning,karimireddy2020SCAFFOLDStochastic} has designed FL algorithms with adjustment of weight decay term in local training. 
However, the impact of weight decay on models can not determine the shape of weight distributions and could not align parameter distributions under different conditions.
Hence, it does not necessarily lead to a harmonization of the model scopes before fusion.

\paragraph{Comparison with Proximal Term.} \label{sec:ana-proximal}
The proximal term is used to keep the model weight $\boldsymbol{w}$ and another one $\tilde{\boldsymbol{w}} \in \mathbb{R}^n$ as close as possible during the update of $\boldsymbol{w}$, which is represented as $||\boldsymbol{w}-\tilde{\boldsymbol{w}}||_2^2$. In FedProx~\citep{li2020FederatedOptimization}, the proximal term constrains the local models to stay close to the global model, thereby ensuring stability during model fusion.
For clarity, we also define $\tilde{\mu}$ and $\tilde{\sigma}$ as the mean and standard deviation of vectors  $\tilde{\boldsymbol{w}}$. 
The proximal term can be decomposed as follows:
\begin{small}
\begin{equation}
    ||\boldsymbol{w}-\tilde{\boldsymbol{w}}||_2^2 = n\sigma^2 + n\mu^2 + n\tilde{\sigma}^2 + n\tilde{\mu}^2 - 2\sum_{i=1}^{n} \boldsymbol{w}_i \tilde{\boldsymbol{w}}_i.
\end{equation}
\end{small}

Because the part of $n\tilde{\sigma}^2 + n\tilde{\mu}^2$ can be seen as a constant term, the proximal term differs from weight decay by only an additional term of $-2\sum_{i=1}^{n} \boldsymbol{w}_i \tilde{\boldsymbol{w}}_i$, which encourage the weights in $\boldsymbol{w}$ to align with the direction of $\tilde{\boldsymbol{w}}$ as closely as possible.
However, the restriction is too strict because it requires the direction alignment for each specific element.
This may limit the normal training process and perturb the effects of other loss functions.

In the context of model fusion, weight decay represents a more flexible approach, while the proximal term seems to introduce directional constraints additionally.
Both of them have overlooked the influence of weight scope on model fusion and are unable to align weight scopes, as demonstrated in \Figref{fig:fig4_std_variation_ws_wd_prox}. 
Our proposed method aims to achieve consistency in model weight ranges for improved model fusion results.

\paragraph{Comparison with Network Invariance.} \label{sec:ana-invariance}
Some studies have investigated the impact of permutation invariance on mode connectivity~\cite{entezari2022RolePermutation,singh2020ModelFusion,ainsworth2023GitRebasin,tatro2020OptimizingMode,ashmore2015MethodFinding,godfrey2022SymmetriesDeep,brea2019WeightspaceSymmetry} and model aggregation in federated learning~\cite{yurochkin2019BayesianNonparametric,li2022FederatedLearning,wang2020FederatedLearning}. 
A more detailed introduction of these studies has been presented in  \Secref{sec:relate}. 
However, the weight scope mismatch could make the alignment algorithm inaccurate, especially the weight-based ones~\cite{singh2020ModelFusion}. 
Additionally, even if the neurons are aligned in order, their weight scopes are still mismatched, which could also lead to performance degradation.
Several works also study the scale invariance, e.g., \cite{pittorino2022DeepNetworks} normalizes the model parameters layer-wisely and then searches the proper neural permutations, and \cite{jordan2023REPAIRREnormalizing} studies the relative scale of weight and bias in monotonic linear interpolation.
None of these works formally studies the impact of weight scope on model fusion.

\section{Applications of WSA} \label{sec:app}

This section presents the applications of Weight Scope Alignment to mode connectivity and federated learning. 
They can be seen as one-stage and multi-stage model fusion scenarios, respectively.
While the specific processes or the model set to be fused may differ across them, they all share the common thread of leveraging the proposed WSA as a constraint during training.

\subsection{Mode Connectivity} \label{sec:app-mc}
Given two well-trained models, $\boldsymbol{w}_{1}$ and $\boldsymbol{w}_{2}$, the loss barrier along their linear interpolation path is:
\begin{small}
\begin{equation}
    \max _{\alpha \in[0,1]}\mathcal{L}\left((1-\alpha) \boldsymbol{w}_1+\alpha \boldsymbol{w}_2\right)-\left[(1-\alpha) \mathcal{L}\left(\boldsymbol{w}_1\right)+\alpha \mathcal{L}\left(\boldsymbol{w}_2\right)\right],
    \label{eq:loss_barrier}
\end{equation}
\end{small}

where $\alpha$ is the interpolation coefficient, and the loss barrier represents the point of maximum loss increase along the linear interpolation path.
A higher loss barrier suggests that the two models may not be in the same basin within the loss landscape, while a lower loss barrier indicates linear mode connectivity.

Both OTFusion~\cite{singh2020ModelFusion} and Git Re-basin~\cite{ainsworth2023GitRebasin} have improved the way they perform model interpolation by considering neuron matching.
They calculate matching relationships $\Pi$ between the weights of each layer in the two models, using a permutation matrix or optimal transport matrix, leading to the fusion formula:
\begin{equation}
(1-\alpha)\boldsymbol{w}_{1} + \alpha \Pi \boldsymbol{w}_{2}.
\end{equation}

Our approach can easily be integrated into the aforementioned model interpolation methods to achieve improved mode connectivity.
For separately trained models, they are typically randomly initialized from the same distribution, and their weight scopes are similar. 
The mean $\tilde{\mu}$ and standard deviation $\tilde{\sigma}$ of the target weight distribution are the hyperparameters and keep invariant during each model's training. We incorporate a Weight Scope Regularization term (i.e., \Eqref{eq:distribution_kl}) to ensure that the weight range remains close to the target weight scope.
With our method, the weight scope of the converged models are matched, which is beneficial for searching the matching matrix.

\subsection{Federated Learning} \label{sec:app-fl}
Model fusion is a fundamental procedure in Federated Learning (FL), i.e., improving the performance of joint training by merging models trained on different data sources.
Its challenge lies in the fact that real-world data distributions across different clients are heterogeneous, leading to a phenomenon known as ``client drift'' during local client training, which impacts the overall performance of FL.
Some recent research has addressed this issue from a Non-I.I.D. perspective, proposing various improved methods.

We use the previously defined $\mathcal{K}$ to represent the set of clients, with each client associated with the data distribution $\mathcal{D}_k$.
The objective of FL is:
\begin{equation}
    \min _{\boldsymbol{w} \in \mathbb{R}^d} f(\boldsymbol{w}), \quad f(\boldsymbol{w}) \triangleq \frac{1}{|\mathcal{K}|} \sum_{k\in \mathcal{K}} \mathcal{L}\left(\boldsymbol{w}; \mathcal{D}_k \right),
\end{equation}
where $\mathcal{L}(\cdot)$ evaluates the loss for each data sample on model $\boldsymbol{w}$.
Cross-entropy is commonly used as the loss function.
To apply our method in FL, we additionally upload the weight scope (i.e., \Eqref{eq:scope-sigma}) to the server after each local training procedure.
The server then performs the fusion of these weight scopes (i.e., \Eqref{eq:distribution_fusion}) and sends the merged weight scopes $\tilde{p}$ back to the clients.
During local client training, the optimization proceeds as follows:
\begin{small}
\begin{equation}
    \label{eq:loss_fedavg_ws}
    \mathcal{L}_{\text{local}}\left(\boldsymbol{w}; \mathcal{D}_k,\tilde{p}\right) = \mathcal{L}\left(\boldsymbol{w}; \mathcal{D}_k \right) + \lambda \sum_{\ell=1}^{L} D_{\text{KL}}(p(\boldsymbol{w}^{\ell}), \tilde{p}^{\ell}),
\end{equation}
\end{small}

where $\lambda$ is used to control the strength of weight scope alignment, and $\tilde{p}^{\ell}$ is the fused weight scope of the $\ell^{th}$ layer.
The full framework of FedAvg~\cite{mcmahan2017CommunicationefficientLearninga} with WSA is outlined in Algorithm 1.

\begin{small}
\begin{algorithm}[tbp]
\caption{FedAvg with \colorM{WSA}}
\begin{algorithmic}[1]
\State \textbf{Input:} model $\boldsymbol{w}$, number of rounds $T$, local iteration steps $\tau$, parameters $\eta$, $\varepsilon$, 
\For{$t = 0, \dots, T-1$ communication rounds}
    \State \textbf{Global server}:
    \State \hspace{\algorithmicindent} Send $\boldsymbol{w}, \colorM{\tilde{p}=\{\tilde{p}^{\ell}=(\tilde{\mu}^{\ell},\tilde{\sigma}^{\ell})\}_{\ell \in [L]}}$ to all clients
    \State Client $k \in \mathcal{K}$ in parallel do:
    \State \hspace{\algorithmicindent} Set $\boldsymbol{w}_k \gets \boldsymbol{w}$
    \State \hspace{\algorithmicindent} \textbf{for} $s = 0, \dots, \tau-1$ local iterations \textbf{do}
        \State \hspace{\algorithmicindent}\hspace{\algorithmicindent} Update $\boldsymbol{w}_k \gets \boldsymbol{w}_k - \eta \nabla \mathcal{L}(\boldsymbol{w}_k;\mathcal{D}_k, \tilde{p})$ \colorM{using loss in \Eqref{eq:loss_fedavg_ws}}
	    \State \hspace{\algorithmicindent}\hspace{\algorithmicindent} Send $\boldsymbol{w}_k, \colorM{\{\mu^{\ell}_k,\sigma^{\ell}_k\}_{\ell \in [L]}}$ to the server
	\State \textbf{Global server}:
	\State \hspace{\algorithmicindent} Update global model $\boldsymbol{w} \gets \frac{1}{|\mathcal{K}|} \sum_{k \in \mathcal{K}} \boldsymbol{w}_k$ 
	\State \hspace{\algorithmicindent} \colorM{Update $\tilde{\mu}^{\ell},\tilde{\sigma}^{\ell}$ using  \Eqref{eq:distribution_fusion}}
\EndFor
\end{algorithmic}
\end{algorithm}
\end{small}

\section{Experiments}
In Section 5.1, we verify that different training conditions lead to variations in weight scope, and the differences between weight scope can affect the performance of model merging. 
In Section 5.2, we explore the effectiveness of the WSA method in the mode connectivity scenario. 
In Section 5.3, we demonstrate the performance improvements achieved by WSA in various federated learning scenarios and analyze the impact of the method.

\subsection{Basic Experiments} \label{sec:obser}

\begin{figure}[htbp]
    \centering
    \includegraphics[width=\linewidth]{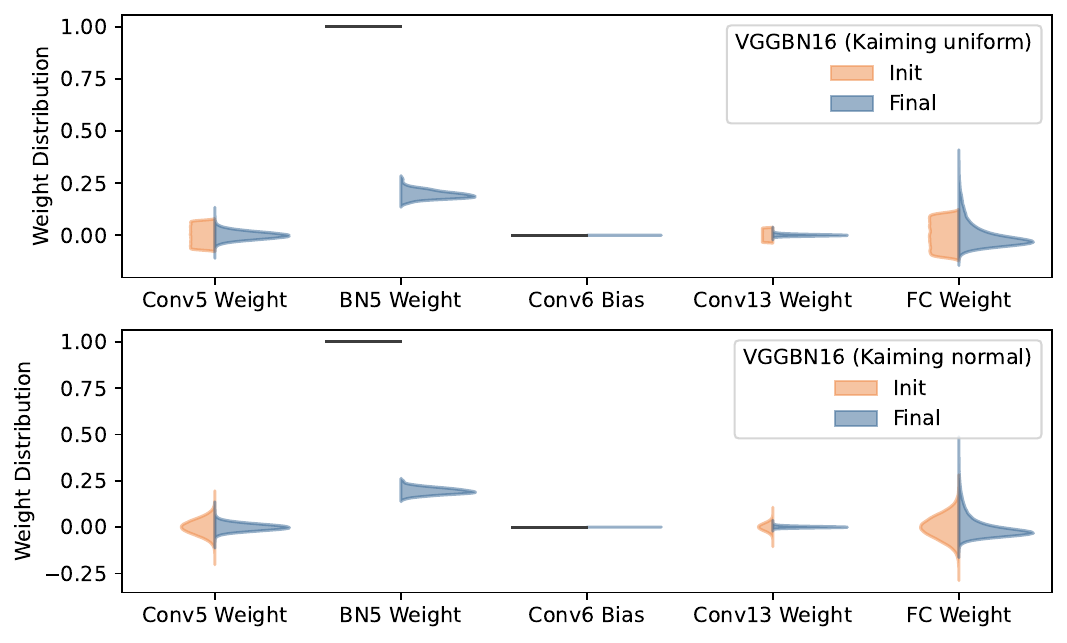}
    \caption{The Gaussian distribution of parameters in VGGBN16 trained on CIFAR-100. Both uniform and normal initialization lead to Gaussian parameter distributions.}
    \label{fig:gaussian-dist}
\end{figure}

\paragraph{Observations.}
Our basic assumption is that the weight scope could be formulated as the Gaussian distribution.
Hence, we first study the weight distributions of the models.
The first observation is that the converged weight scope is irrelevant to the way of weight initialization, e.g., the Kaiming uniform or the Kaiming normal initialization method~\cite{he2015DelvingDeep}.
\Figref{fig:gaussian-dist} shows the weight distributions of several layers in VGG16~\cite{simonyan2015VeryDeep} with BatchNorm~\cite{ioffe2015BatchNormalization}, where we train the network on CIFAR-100~\cite{krizhevsky2009LearningMultiple} for 200 epochs.
In the Appendix, we will present the weight scopes in pre-trained models, which also follow the Gaussian distributions.

\paragraph{The Influence of Weight Scope on Model Fusion.}
We then investigate the difference of weight scope under different conditions, which include: 1) hyperparameters, e.g., optimizer, batch-size, learning rate, and weight decay; 2) data quality, e.g., label imbalance, label noise, feature noise, and data size.
For each condition, we select two specific settings, and then train models under the same setting or not.
For example, the optimizer could be SGD or Adam~\cite{kingma2015AdamMethod}, and we train three models with each using SGD, SGD, and Adam as the optimizer, respectively.
Then, we plot the weight scopes of these models and investigate the linear interpolation accuracy.
\Figref{fig:scope-mc-demo} shows that models trained using different optimizers own varying weight scopes and the interpolation meets an obvious barrier.
The details of training conditions and the corresponding illustration results can be found in the Appendix.

\begin{table}[tbp]
\centering
\caption{The impact of training conditions on LMC.}
\label{tab:factor-barrier}
\resizebox{\columnwidth}{!}{
    \begin{tabular}{@{}lcccc@{}}
    \toprule
    Factor & KL D. {\tiny ($\times 10^{-4}$)} & Ba$(=)$ {\tiny (\%)} & Ba$(\neq)$ {\tiny (\%)} & Diff {\tiny (\%)}  \\ 
    \midrule \midrule
    Label Imb. & 9.60 & 4.60 & 22.05 & 17.45 \\
    Optimizer & 8.95 & 1.13 & 17.84 & 16.71 \\
    Batch Size & 5.96 & 44.98 & 56.07 & 11.09 \\
    Label Noise & 1.49 & 0.58 & 0.79 & 0.21 \\
    Learning Rate & 0.67 & 0.38 & 6.77 & 6.39 \\
    Feature Noise & 0.60 & 0.70 & 9.60 & 8.90 \\
    Data Size & 0.37 & 0.00 & 0.75 & 0.75 \\
    Weight Decay & 0.31 & 1.59 & 2.31 & 0.72 \\
    \bottomrule
    \end{tabular}
}
\end{table}

From \Figref{fig:scope-mc-demo}, we can observe that the weight scopes under the same training condition are near the same.
In fact, we calculate their KL divergence, and the results are near zero.
Hence, we calculate the average KL divergence of weight scopes under different conditions, e.g., the average divergence of Gaussian distributions in the bottom five sub-figures (the ``KL D.'' column in \Tabref{tab:factor-barrier}).
Additionally, we calculate the barriers of the two interpolated curves, and their difference, which are listed in columns of ``Ba$(=)$'', ``Ba$(\neq)$'', and ``Diff'' in \Tabref{tab:factor-barrier}.
In the table, the ``Ba$(=)$'' column is commonly smaller than ``Ba$(\neq)$'', which verifies that models under the same condition are indeed similar in weight scopes.
Empirically, for models under different conditions, a larger KL divergence corresponds to a larger barrier.
This shows that the weight scope mismatch is strongly correlated with the model fusion performance.

\begin{figure}[htbp]
    \centering
    \includegraphics[width=\linewidth]{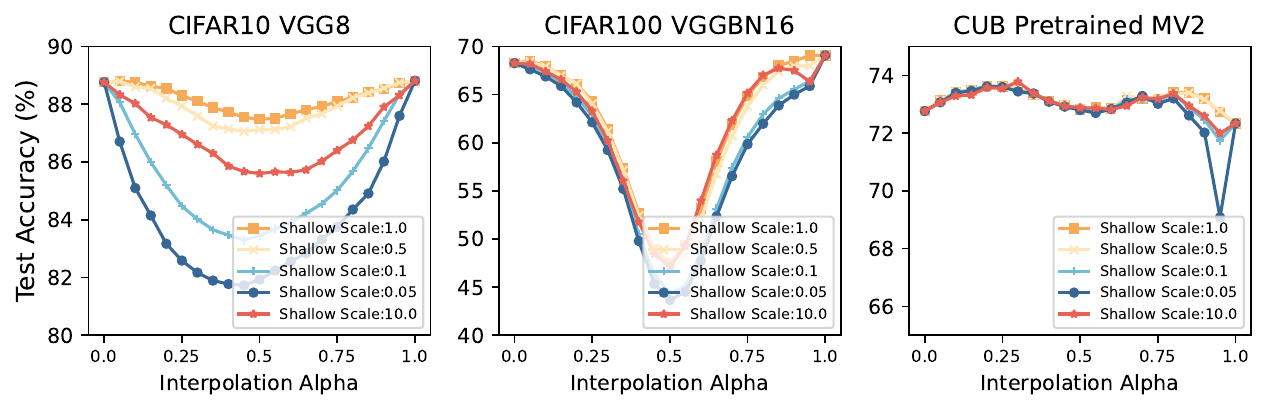}
    \caption{The linear interpolation curves between a model with another model (Scale=1.0) and its scaled versions (Scale $\neq$ 1.0).}
    \label{fig:manual-scale}
\end{figure}

\paragraph{Model Fusion under Manual Scaling.}
The above experiments could not support that the weight scope is the cause of the barrier in model fusion.
Thanks to the scale invariance of neural networks, we could manually scale the networks and create the weight scope mismatch phenomenon.
Specifically, we train two networks under the same condition with different random initialization.
To avoid overuse of symbols, we denote them as $\boldsymbol{w}_1$ and $\boldsymbol{w}_2$, respectively.
To create the scope mismatch, we select one layer in $\boldsymbol{w}_2$ and multiply its weight by $\alpha$, and divide its following layer's weight by $\alpha$.
The obtained model is denoted as $\boldsymbol{w}_2.\alpha$, which performs the same as $\boldsymbol{w}_2$.
However, when taking interpolations  between $\boldsymbol{w}_1$ and $\boldsymbol{w}_2.\alpha$, the interpolation curves differ a lot.
The results are shown in \Figref{fig:manual-scale}, where the ``Scale'' in the legend denotes $\alpha$.
Using $\alpha=1.0$ means no scaling.
Clearly, scaling the layers could make the barrier more obvious, especially in VGG8.
Experimental details and more analysis can be found in the Appendix.

\subsection{Performance in Mode Connectivity}
\begin{figure}[tb]
    \centering
    \includegraphics[width=0.8\linewidth]{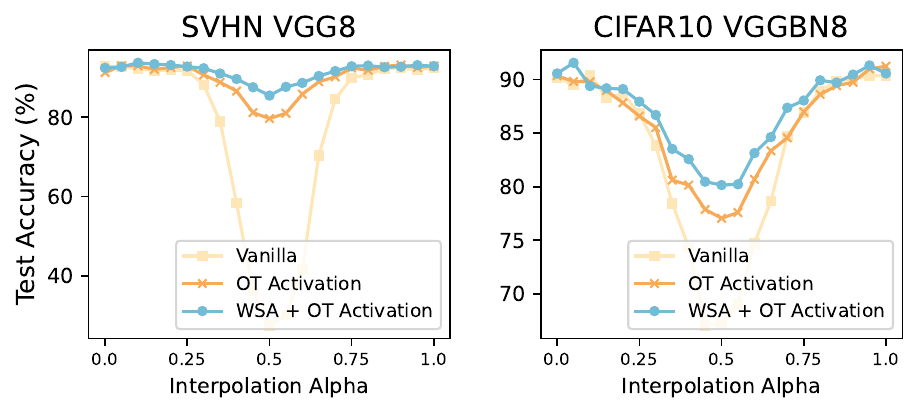}
    \caption{Model interpolation on SVHN and CIFAR-10. WSA enhances OTFusion~\cite{singh2020ModelFusion} performance.}
    \label{fig:ot-ws}
\end{figure}

To investigate the effectiveness of our approach on mode connectivity, we apply the proposed WSA to OTFusion~\cite{singh2020ModelFusion} and Git Re-Basin~\cite{ainsworth2023GitRebasin}. 
Specifically, we first initialize and train two models, and plot the vanilla interpolation curve.
Then, we use the activation-based OTFusion method to search for an alignment and plot the interpolated curve after matching. 
Finally, we replace the models with another two models trained using the WSA as introduced in \Secref{sec:app-mc} and also use the OTFusion to search the alignment matrix. 
The results are shown in \Figref{fig:ot-ws}. 
OTFusion could decrease the barrier of vanilla interpolation, and our proposed WSA could improve its performance further.

\begin{figure}[tbp]
    \centering
    \includegraphics[width=\columnwidth]{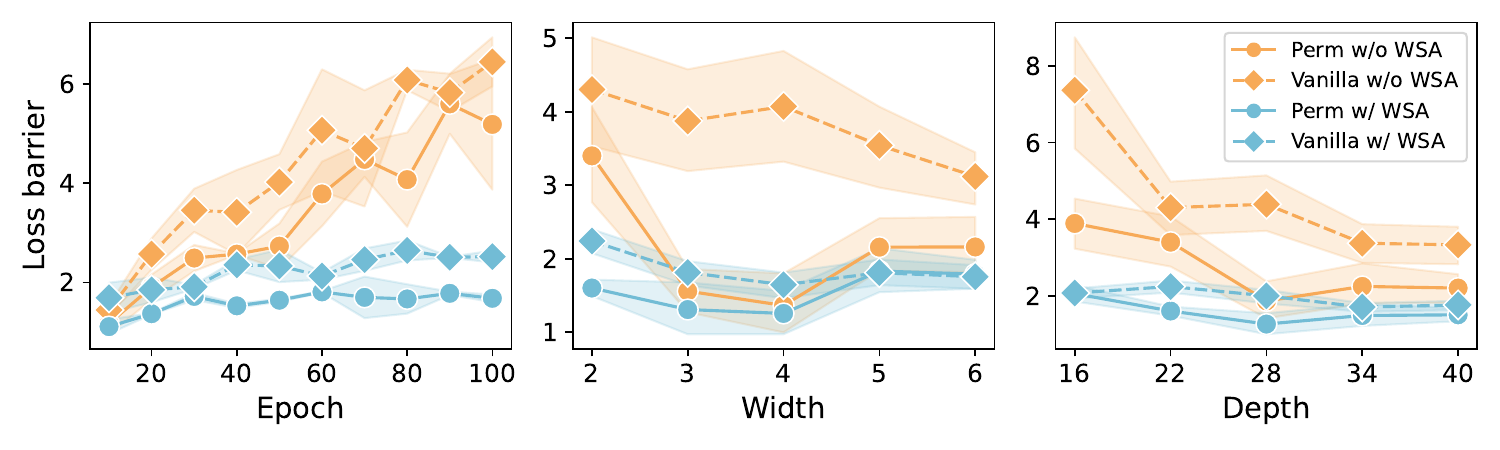}
    \caption{WSA facilitates Git Re-Basin~\cite{ainsworth2023GitRebasin} under different epochs, widths and depths, ensuring a smaller loss barrier.}
    \label{fig:fig_lmc_resnet_width_depth}
\end{figure}

Then, we combine our WSA with Git Re-Basin, which searches the permutation matrix instead of an optimal transport matrix compared with OTFusion. 
Similar to OTFusion, we also train models with or without our proposed WSA.
In \Figref{fig:fig_lmc_resnet_width_depth}, we conduct experiments on a ResNet-22x2~\cite{he2016DeepResidual,ainsworth2023GitRebasin} model on CIFAR-10~\cite{krizhevsky2009LearningMultiple} to compare the loss barrier at different epochs.
The solid lines represent the vanilla model interpolation, while the dashed lines represent the interpolation after using Git Re-basin.
On the one hand, applying Git Re-Basin could indeed decrease the barrier of vanilla interpolation.
However, as the number of epochs increases, the results of w/o WSA continue to grow, even when permutation is considered.
In contrast, the lines of w/ WSA can consistently maintain lower loss barriers. 
In the middle and right figures, we compare loss barriers under different model widths and depths. 
Specifically, we compare ResNet-22 models with a widening factor of 2, 4, and 6, and ResNet models with a widening factor of 2 and depths of 16, 22, and 28.
The results show that our method consistently improves model fusion across models with varying depths and widths.
Furthermore, permutation interpolation further improves the loss barrier, demonstrating that considering both weight permutation and scope can effectively reduce the loss barrier.
This indicates that WSA helps align the weight ranges of models, contributing to enhanced mode connectivity.

\begin{figure}[tbp]
    \centering
    \includegraphics[width=0.7\linewidth]{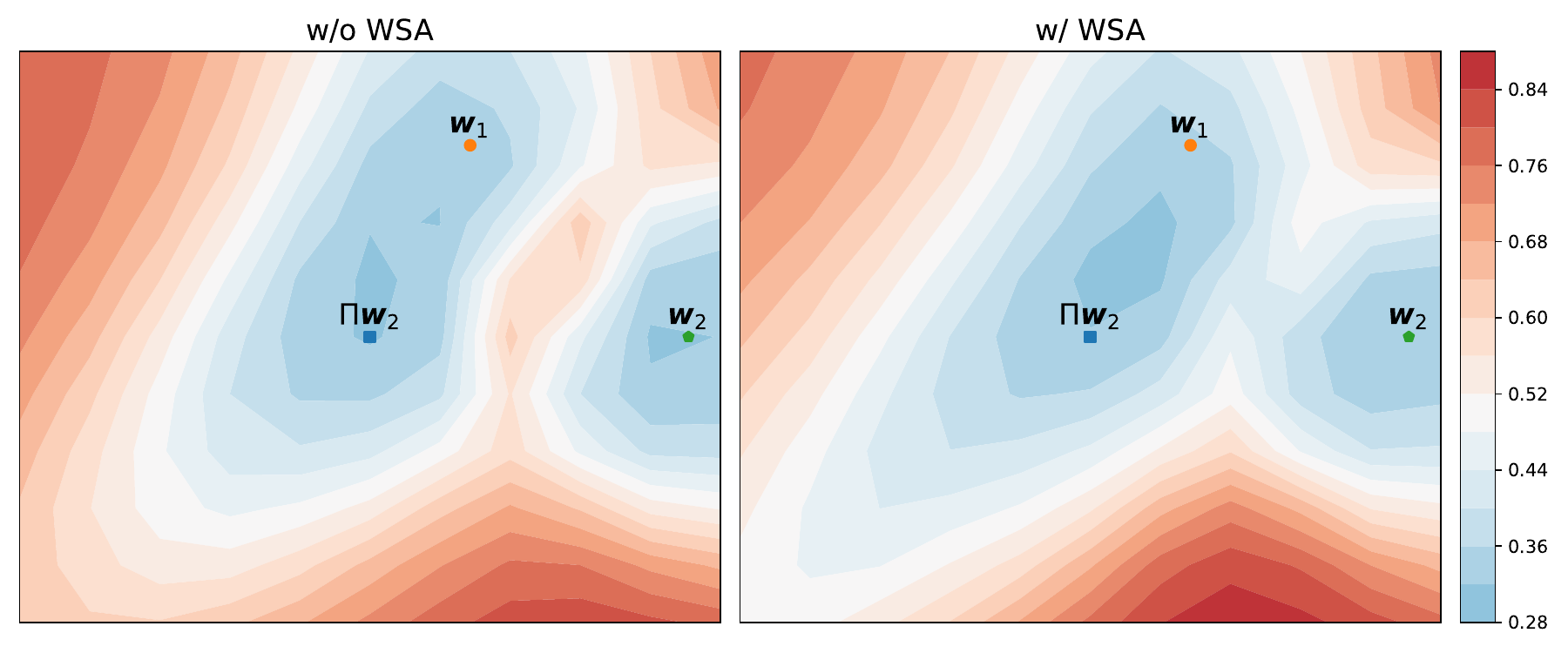}
    \caption{Loss landscape w/o WSA and w/ WSA.}
    \label{fig:fig_lmc_2d_loss_landscape}
\end{figure}

In \Figref{fig:fig_lmc_2d_loss_landscape}, we illustrate the loss landscape of model $\boldsymbol{w}_1,\boldsymbol{w}_2$ and $\Pi\boldsymbol{w}_2$, where the permutation matrix $\Pi$ is computed by Git Re-basin. We train the two models with and without our proposed WSA, separately. First, there is an obvious peak between $\boldsymbol{w}_1$ and $\boldsymbol{w}_2$ on the left; meanwhile, on the right, the loss between $\boldsymbol{w}_1$ and $\boldsymbol{w}_2$ is alleviated with WSA. Secondly, through Git Re-basin, the valley between $\Pi\boldsymbol{w}_2$ and $\boldsymbol{w}_1$ is a larger flat area which leads to better model merging.

\begin{figure}[htp]
    \centering
    \begin{minipage}[b]{0.23\textwidth}
        \centering
        \includegraphics[width=\textwidth]{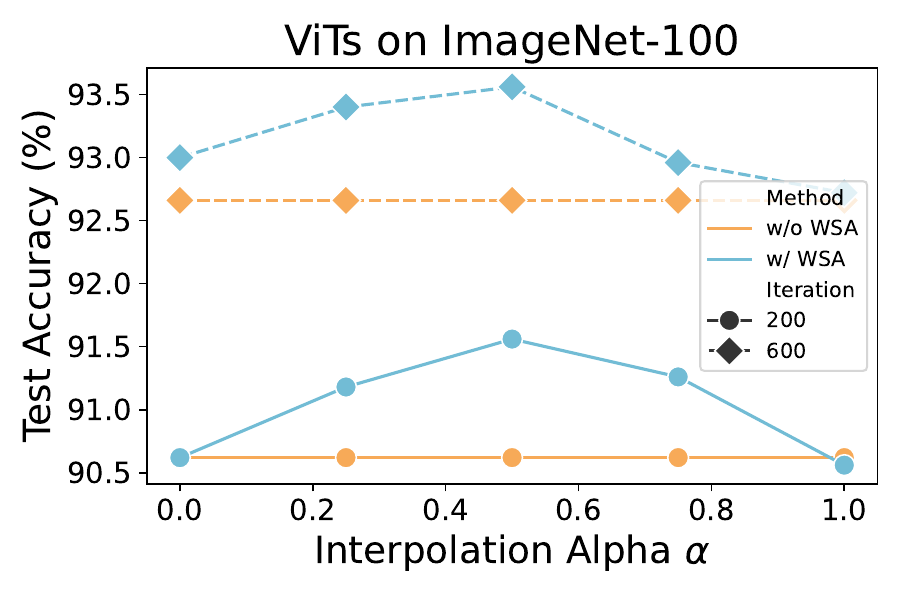}
        \caption{Mode Connectivity of ViTs finetuned on ImageNet-100.}
        \label{fig:fig8_pretrained_lmc}
    \end{minipage}
    \hfill
    \begin{minipage}[b]{0.23\textwidth}
        \centering
        \includegraphics[width=\textwidth]{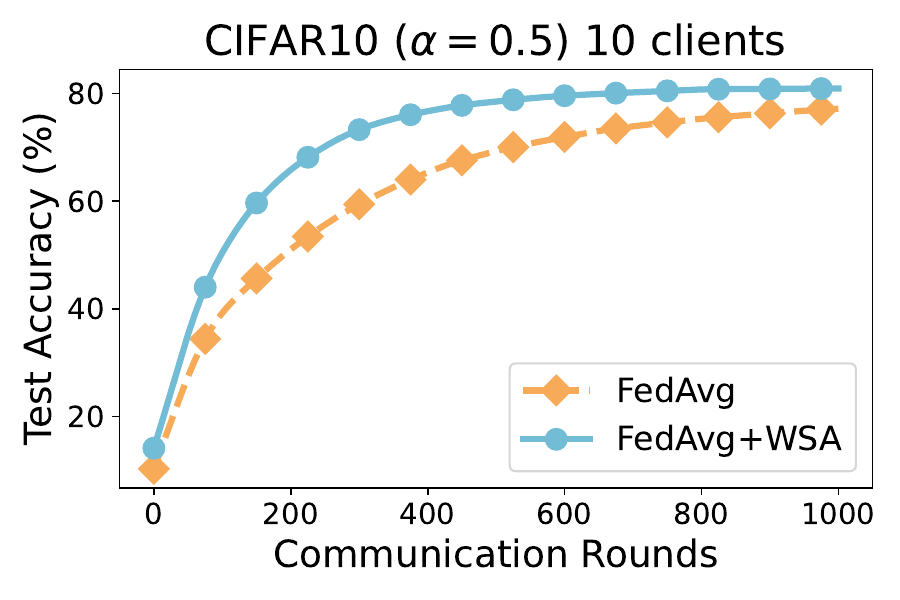}
        \caption{Test accuracy curve of VGG8 trained on CIFAR-10 in FL.}
        \label{fig:fig6_convergence_vgg8}
    \end{minipage}
\end{figure}

\paragraph{Pretrained Model.} We conduct experiments using the Vision Transformer (ViT) pre-trained on ImageNet-21k on mode connectivity.
\Figref{fig:fig8_pretrained_lmc} presents the results of finetuning the pre-trained ViT on ImageNet-100. 
Specifically, We finetune the ViT using two distinct random seeds and then conduct model interpolation between the resultant models.
``baseline'' refers to training without intervation, while ``w/ WSA'' indicates training with Weight Scope Alignment.
Because of permutation-based weight averaging methods are not designed for transfomers, we employ the naive model interpolation $(1-\alpha)\boldsymbol{w}_{1} + \alpha\boldsymbol{w}_{2}$. 
The results demonstrate that WSA can improve the perfromance of interpolated models.

\subsection{Performance in Federated Learning}
To simulate real-world federated learning (FL) scenarios, we initially consider image classification tasks across three datasets: CIFAR-10~\cite{krizhevsky2009LearningMultiple}, CIFAR-100~\cite{krizhevsky2009LearningMultiple}, and CINIC-10~\cite{darlow2018CINIC10Not}. 
We explore two different client scenarios: ``cross-device'' with 100 small-data clients where only 10 clients participate in each training round, and ``cross-silo'' with 10 clients, each having larger datasets and engaging each training round.
We use the same CNN architecture used in \cite{karimireddy2020SCAFFOLDStochastic,jhunjhunwala2023FedExPSpeeding}.

To ensure Non-I.I.D. data distribution in data partitioning, we employ the Dirichlet distribution for creating heterogeneous data on each client~\cite{yu2021Fed2Featurealigned,li2022FederatedLearning}.
We compare with several state-of-the-art (SOTA) methods, including FedAvg~\cite{mcmahan2017CommunicationefficientLearninga}, FedProx~\cite{li2020FederatedOptimization}, SCAFFOLD~\cite{karimireddy2020SCAFFOLDStochastic}, and FedExp~\cite{jhunjhunwala2023FedExPSpeeding}. 
Throughout all experiments, unless otherwise specified, we use a default batch size of 50, 20 local update steps, and 1500 communication rounds.
Additional details and results are provided in the appendix.

\begin{table*}[tbp]
    \centering
    \caption{Comparison of Test Accuracy (\%) across various FL settings on three datasets, involving 10 clients: ``Baseline'' represents the original method, while ``+WSA'' indicates the incorporation of WSA into ``Baseline''. Alpha $\downarrow$ denotes client heterogeneity $\uparrow$. \textbf{Bold} fonts highlight the optimal results between ``Baseline'' and ``+WSA''. The results demonstrate that +WSA enhances performance in all FL settings.}
    \label{tab:fed_client10_baseline}
    \resizebox{\textwidth}{!}{
        \begin{tabular}{@{}lcccccccccccc@{}}
        \toprule
        \multicolumn{1}{l|}{Dataset} & \multicolumn{4}{c|}{CIFAR-10} & \multicolumn{4}{c|}{CIFAR-100} & \multicolumn{4}{c}{CINIC-10} \\ \cmidrule(l){2-13} 
        \multicolumn{1}{l|}{Alpha} & \multicolumn{2}{c}{0.5} & \multicolumn{2}{c|}{1} & \multicolumn{2}{c}{0.5} & \multicolumn{2}{c|}{1} & \multicolumn{2}{c}{0.5} & \multicolumn{2}{c}{1} \\ \cmidrule(l){2-13} 
        \multicolumn{1}{l|}{Algorithm} & Baseline & +WSA & Baseline & \multicolumn{1}{c|}{+WSA} & Baseline & +WSA & Baseline & \multicolumn{1}{c|}{+WSA} & Baseline & +WSA & Baseline & +WSA \\ 
        \midrule \midrule
        FedAvg~\cite{mcmahan2017CommunicationefficientLearninga} & 68.42 & \textbf{72.39} & 69.06 & \textbf{73.09} & 30.66 & \textbf{35.23} & 30.86 & \textbf{36.16} & 53.05 & \textbf{57.88} & 53.94 & \textbf{58.53} \\
        FedProx~\cite{li2020FederatedOptimization} & 68.40 & \textbf{72.49} & 69.11 & \textbf{73.18} & 30.47 & \textbf{35.39} & 31.20 & \textbf{36.24} & 52.57 & \textbf{57.54} & 53.34 & \textbf{58.26} \\
        SCAFFOLD~\cite{karimireddy2020SCAFFOLDStochastic} & 66.69 & \textbf{71.61} & 68.14 & \textbf{72.65} & 29.58 & \textbf{34.38} & 31.12 & \textbf{36.76} & 50.88 & \textbf{55.87} & 52.13 & \textbf{57.52} \\
        FedExP~\cite{jhunjhunwala2023FedExPSpeeding} & 73.24 & \textbf{75.55} & 72.73 & \textbf{76.10} & 38.93 & \textbf{41.71} & 40.48 & \textbf{43.16} & 56.62 & \textbf{60.77} & 56.30 & \textbf{59.93} \\ \bottomrule
        \end{tabular}
    }
\end{table*}

\begin{table*}[]
    \centering
    \caption{Comparison of Test Accuracy (\%) across various FL settings on three datasets, involving 100 clients where 10\% participate in each round: ``Baseline'' represents the original method, while ``+WSA'' indicates the incorporation of WSA into ``Baseline''. Alpha $\downarrow$ denotes client heterogeneity $\uparrow$. \textbf{Bold} fonts highlight the optimal results between ``Baseline'' and ``+WSA''. The results demonstrate that +WSA enhances performance in all FL settings.}
    \label{tab:fed_client100_baseline}
    \resizebox{\textwidth}{!}{
    \begin{tabular}{lcccccccccccc}
        \toprule
        \multicolumn{1}{l|}{Dataset} & \multicolumn{4}{c|}{CIFAR-10} & \multicolumn{4}{c|}{CIFAR-100} & \multicolumn{4}{c}{CINIC-10} \\ \cmidrule(l){2-13} 
        \multicolumn{1}{l|}{Alpha} & \multicolumn{2}{c}{0.5} & \multicolumn{2}{c|}{1} & \multicolumn{2}{c}{0.5} & \multicolumn{2}{c|}{1} & \multicolumn{2}{c}{0.5} & \multicolumn{2}{c}{1} \\ \cmidrule(l){2-13} 
        \multicolumn{1}{l|}{Algorithm} & Baseline & +WSA & Baseline & \multicolumn{1}{c|}{+WSA} & Baseline & +WSA & Baseline & \multicolumn{1}{c|}{+WSA} & Baseline & +WSA & Baseline & +WSA \\ 
        \midrule \midrule
        FedAvg & 66.77 & \textbf{71.11} & 69.06 & \textbf{72.80} & 29.57 & \textbf{34.47} & 30.45 & \textbf{35.31} & 50.85 & \textbf{56.68} & 52.64 & \textbf{57.82} \\
        FedProx & 66.82 & \textbf{71.14} & 69.06 & \textbf{72.73} & 29.72 & \textbf{34.41} & 30.35 & \textbf{35.17} & 50.14 & \textbf{56.31} & 52.13 & \textbf{57.55} \\
        SCAFFOLD & 65.74 & \textbf{70.36} & 69.47 & \textbf{73.25} & 29.37 & \textbf{34.85} & 31.81 & \textbf{37.31} & 48.39 & \textbf{54.49} & 52.84 & \textbf{57.07} \\
        FedExP & 71.79 & \textbf{74.33} & 71.82 & \textbf{74.80} & 37.98 & \textbf{41.34} & 38.46 & \textbf{41.58} & 50.22 & \textbf{56.99} & 55.84 & \textbf{60.55} \\ 
        \bottomrule
    \end{tabular}
    }
\end{table*}

\Tabref{tab:fed_client10_baseline} presents the performance comparison of CNN model trained in various FL settings with 10 clients where FedAvg+WSA outperforms FedAvg in performance. 
Moreover, as a plug-in, we can easily adapt it to existing FL methods, and the experiments indicate that incorporating WSA leads to significant improvements in each method. 
This underscores the effectiveness of our approach in model fusion of FL. 
\Tabref{tab:fed_client100_baseline} shows that similar results are observed with 100 clients.
Furthermore, \Figref{fig:fig6_convergence_vgg8} illustrates the adaptability of WSA to deeper models on CIFAR-10.

\begin{figure}[tbp]
    \centering
    \includegraphics[width=\columnwidth]{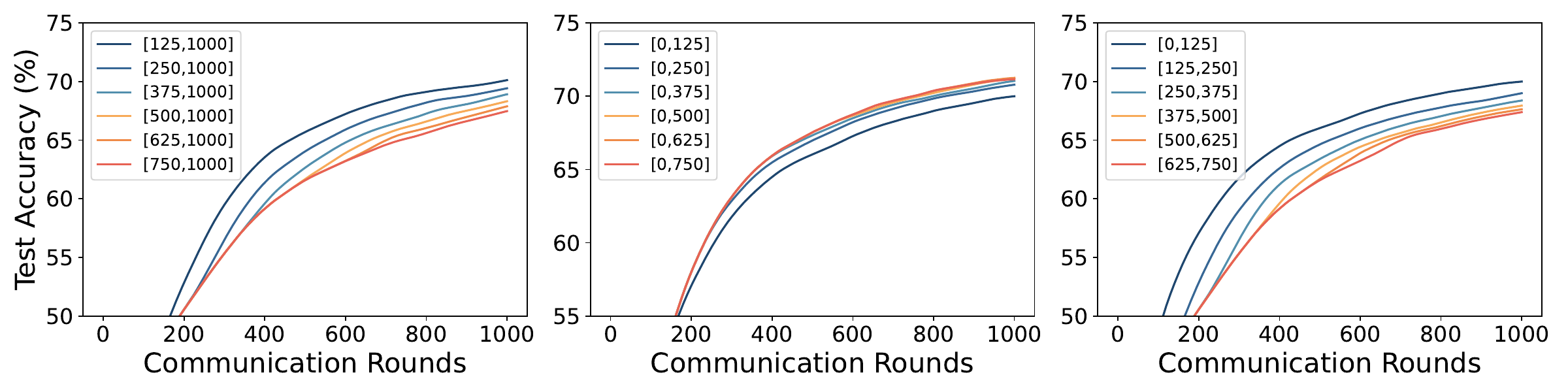}
    \caption{The Impact of using WSA in different rounds in FL.}
    \label{fig:different_influence_rounds}
\end{figure}

To further understand the impact of WSA on model merging, we conduct experiments on the CIFAR-10 dataset  over various durations.
We consider three different intervention times during training: from the start of training until iteration $[0,t_1]$, from iteration $[t_1,T]$ until convergence, and during the iterations within the interval $[t_1, t_2]$.
\Figref{fig:different_influence_rounds} displays the convergence curves for these three scenarios.
In the first scenario, we observe that the longer the intervention time, the better our method enhances the convergence performance.
In the second one, we find that employing our method earlier results in greater performance improvements and faster convergence.
Lastly, even if our method only intervenes during the initial $[0, 125]$ rounds, it still significantly boosts performance.
Drawing inspiration from the concept of a ``critical learning period''~\cite{yan2021CriticalLearning,achille2019CriticalLearning}, we conclude that assisting weight alignment in the early stages of learning provides more substantial benefits for the model fusion scenario.

\begin{figure}[tbp]
    \centering
    \includegraphics[width=\columnwidth]{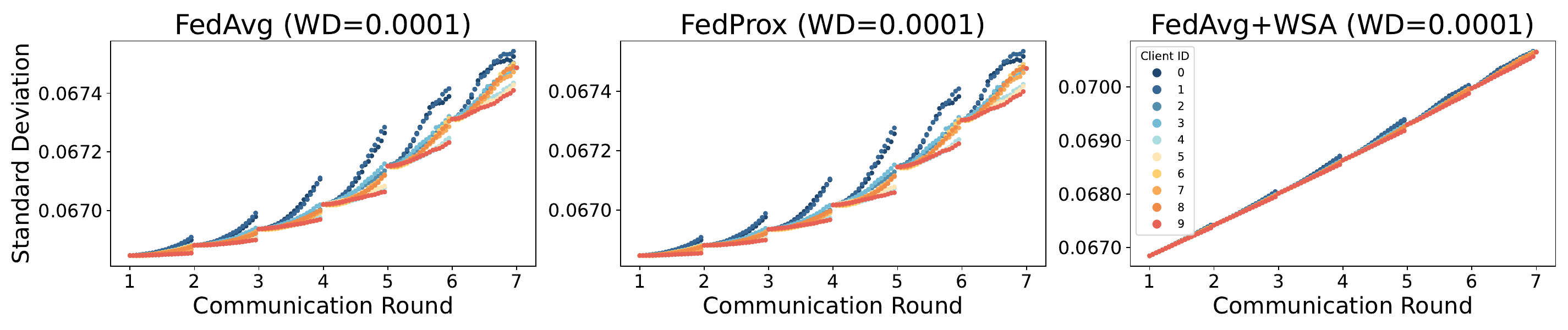}
    \caption{The standard deviation of conv2 weights for each client in FL training with 10 clients, varying across rounds.}
    \label{fig:fig4_std_variation_ws_wd_prox}
\end{figure}

\paragraph{Analysis of weight distribution.} In \Figref{fig:fig4_std_variation_ws_wd_prox}, we illustrate the alignment effect introduced by WSA on parameter ranges during the training process of FedAvg. 
The figure displays how the weight distributions of 10 clients evolve during training. 
To enhance clarity, we represent the distributions separately using mean and variance, showing the results for the original FedAvg, FedProx, and FedAvg+WSA.
Different colors represent different clients, and aggregation occurs when the server averages the model parameters across all clients after each communication round, making the model distributions identical at that point.
In the variance plots, we observe aggregation and divergence in FedAvg even with a weight decay of 0.0001, particularly in the early stages, indicating that the mismatch of weight ranges may contribute to slow convergence in the early phases of FL. 
In contrast, our constraint promotes a more cohesive model variance. 

\begin{figure}[htbp]
    \centering
    \includegraphics[width=0.7\linewidth]{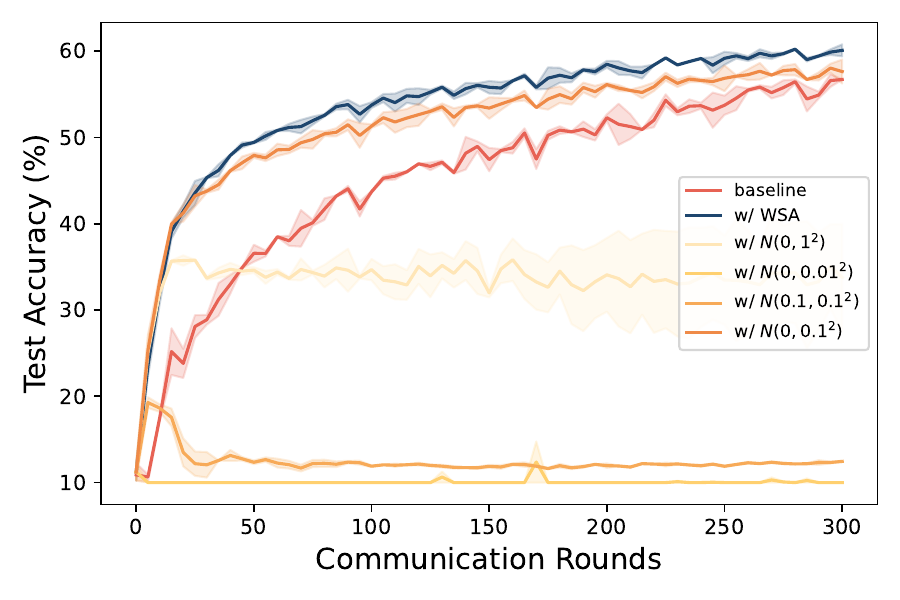}
    \caption{Comparison with pre-defined distributions in FL.}
    \label{fig:fig7_predefine_in_FL}
\end{figure}

\paragraph{Comparison with pre-defined distributions.} To demonstrate the effectiveness of weight scope fusion, we try to design some pre-defined distributions to replace it. 
Given that Gaussian distributions match the frequently noted attributes of model weights, we devise various pre-defined Gaussian distributions $N(\mu, \sigma^2)$, by altering the mean $\mu$ and standard deviation $\sigma$. 
Each of these distributions is utilized to substitute the fusion distributions in WSA, and are set across all layers.
As shown in \Figref{fig:fig7_predefine_in_FL}, our method ``w/ WSA'' outperforms all other comparative approaches. 
The pre-defined distribution w/ $N(0,0.1^2)$ is the second-best result, while other pre-defined ones led to a degradation in performance.
We believe that a well-designed fix distribution can be beneficial during the initial training stage. 
Yet, weight distribution variation significantly shifts throughout the training, as seen in \Figref{fig:fig4_std_variation_ws_wd_prox}, necessitating a dynamic and adaptable fusion distribution.
Hence, our suggested weight scope fusion becomes crucial for broad model fusion applications.

\section{Conclusion}
In this study, we investigate the impact of different weight ranges of model parameters on model merging.
We observe that models trained under various conditions exhibit inconsistencies in their weight ranges, leading to a decrease in fusion performance.
To address this issue, we propose Weight Scope Alignment method, which utilize Weight Scope Regularization to constrain the alignment of weight scopes during training, ensuring that the model's weights closely match the specified scope. 
Moverover, for multi-stage fusion scenarios, we design Weight Scope Fusion to fuse the weight scopes of multiple models and regards the fused one as the target weight scope for Weight Scope Regularization.
This alignment enhances performance during model merging.
We validate the effectiveness of our approach in mode connectivity and federated learning.

\begin{ack}
This research was supported by National Science and Technology Major Project (2022ZD0114805), NSFC (61773198, 62376118,61921006), Collaborative Innovation Center of Novel Software Technology and Industrialization.
\end{ack}

\bibliography{1_var}

\clearpage
\setcounter{page}{1}

\section{Dataset and Network Details}

\subsection{Datasets}
In this paper, we use SVHN~\citep{netzer2011ReadingDigits}, CIFAR-10~\citep{krizhevsky2009LearningMultiple}, CIFAR-100~\citep{krizhevsky2009LearningMultiple}, CINIC-10~\citep{darlow2018CINIC10Not} and CUB~\citep{wah2011CaltechucsdBirds2002011} datasets. We provide a general introduction to these datasets as follows:
\begin{enumerate}
    \item SVHN~\citep{netzer2011ReadingDigits}: The Street View House Numbers (SVHN) dataset, a real-world image collection used in \Figref{fig:ot-ws}, features small, cropped digits captured from street view images, often exhibiting variations in rotation. It comprises 73,257 digit images for training and 26,032 for testing, spanning across 10 classes.
    \item CIFAR-10~\citep{krizhevsky2009LearningMultiple}: The CIFAR-10 dataset comprises 60,000 32x32 color images, evenly distributed across 10 classes with each class containing 6,000 images.
    \item CIFAR-100~\citep{krizhevsky2009LearningMultiple}: The CIFAR-100 dataset consists of 100 classes, each with 600 images. These classes are further divided into 20 superclasses, grouping similar categories together.
    \item CINIC-10~\citep{darlow2018CINIC10Not}: The CINIC-10 dataset, containing a total of 270,000 images, is derived from two sources: ImageNet and CIFAR-10. It is divided into three subsets: training, validation, and testing, with each subset comprising 90,000 images. In our study, the validation subset is not utilized.
    \item CUB~\citep{wah2011CaltechucsdBirds2002011}: The Caltech-UCSD Birds-200-2011 (CUB) dataset is designed for fine-grained visual categorization tasks, featuring 11,788 images across 200 bird subcategories. It is divided into 5,944 training images and 5,794 testing images.
\end{enumerate}

\subsection{Networks}
In the work, our used networks include VGG~\citep{simonyan2015VeryDeep}, ResNet~\citep{he2016DeepResidual}, Wide ResNet~\citep{zagoruyko2016WideResidual}, ResNeXt~\citep{xie2017AggregatedResidual}, MobileNet-v2~\citep{howard2017MobileNetsEfficient}. Here is a general introduction to these networks as follows:
\begin{enumerate}
    \item CNN: The model architecture consists of two convolutional layers, the first with 32 $5 \times 5$ filters and the second with 64 $5 \times 5$ filters, followed by two linear layers containing 384 and 194 neurons, respectively. A softmax layer concludes the architecture. This configuration is commonly utilized in studies \citep{mcmahan2017CommunicationefficientLearninga, jhunjhunwala2023FedExPSpeeding, karimireddy2020SCAFFOLDStochastic}.
    \item VGG~\citep{simonyan2015VeryDeep}: VGG models consist of multiple convolution layers and fully-connected layers for classification, typically with 11, 13, 16, or 19 layers. We do not employ batch normalization in VGG11, VGG13, VGG16, or VGG19 (referred to as VGG11-BN, VGG13-BN, VGG16-BN, and VGG19-BN when batch normalization is used). Additionally, we utilize 8-layer and 9-layer VGG networks without batch normalization (VGG8 and VGG9), as mentioned in \citep{yu2021Fed2Featurealigned, li2022FederatedLearning}.
    \item ResNet~\citep{he2016DeepResidual}: ResNet, leveraging residual connections and batch normalization, aims for enhanced performance and robustness. We use ResNet-18 and ResNet-32 in this work. The implementation code we use is provided by \citep{jhunjhunwala2023FedExPSpeeding}.
    \item Wide ResNet~\citep{zagoruyko2016WideResidual}: Wide ResNet (WRN), an extension of ResNet, varies in width to offer different capacities. Due to increased training time with wider networks, we explore variations like WRN16x2, 22x2, 28x2, 34x2, 40x2, and 22x3, 22x4, 22x5, 22x6.
    \item ResNeXt~\citep{xie2017AggregatedResidual}: ResNeXt substitutes group convolutions for partial convolutions in ResNet. We utilize the pre-trained ResNeXt models available in PyTorch for analyzing weight scope distributions, as they offer aggregated residual transformations.
    \item MobileNet-v2~\citep{howard2017MobileNetsEfficient}: MobileNet-v2, known for its compact structure suitable for portable devices, is employed in our study using its pre-trained version available in PyTorch to analyze weight scope distributions.
\end{enumerate}

We will further introduce the datasets and networks used in each application. All our experiments are conducted on a NVIDIA 3090Ti GPU.

\section{Experimental Details}
\subsection{Weight Scope Mismatch}
We first explain the experimental setup for the phenomenon of weight scope mismatch in \Figref{fig:scope-mc-demo}, which the influenced factor is the type of optimizer. 
Specifically, \Figref{fig:scope-mc-demo} trains VGG8 network on CIFAR-10 dataset with 200 epochs. 
We first use SGD optimizer to individually train two models, which are named M0 and M1, respectively. 
Then we use Adam optimizer to train model M2. 
We select the parameters including ``conv1.weight'', ``conv3.weight'', ``conv3.bias'', ``conv5.weight'', and ``fc.weight'' and show their weight distributions. 
Other hyperparameters are listed as follows: the weight decay is $10^{-5}$, the batch size is 128, the learning rate is 0.03 for SGD optimizer, while the learning rate is 0.0003 for Adam. 
The momentum for SGD is 0.9 by default. 
The weight distributions are visualized by seaborn~\footnote{\url{https://seaborn.pydata.org/generated/seaborn.violinplot.html}}.

Furthermore, we explore the following key factors that may influence the weight scope of parameters:
\begin{enumerate}
    \item \textbf{Learning Rate}: we use 0.03 for M0 and M1, while we use 0.001 for M2.
    \item \textbf{Weight Decay}: we use $10^{-5}$ for M0 and M1, while we use $10^{-3}$ for M2.
    \item \textbf{Batch Size}: we use 32 for M0 and M1, while we use 1024 for M2.
    \item \textbf{Optimizer}: we use SGD for M0 and M1, while we use Adam for M2.
    \item \textbf{Feature Noise}: we add gaussian noise to the input features, sampled from the gaussian distribution $\mathcal{N}(1.0, \epsilon^2)$. We set $\epsilon=0.0$ for M0 and M1, while we set $\epsilon=0.1$ for M2.
    \item \textbf{Label Noise}: the label of each training sample is flipped to a random class with a probability of $p$. We set $p=0.0$ for M0 and M1, indicating no label flipped. We set $p=0.1$ for M2.
    \item \textbf{Dataset Size}: a fraction $q$ of data is randomly selected  to train the model. We set $q=1.0$ for M0 and M1, while we set $q=0.1$ for M2.
    \item \textbf{Label Imbalance}: we split the training data into two parts. The first part contains $90\%$ samples of the first five classes and $10\%$ samples of the last five classes, while the second part contains other samples. We train M0, M1 on the first part, while we train M2 on the second part.
\end{enumerate}

\begin{figure}[tb]
    \centering
    \includegraphics[width=\linewidth]{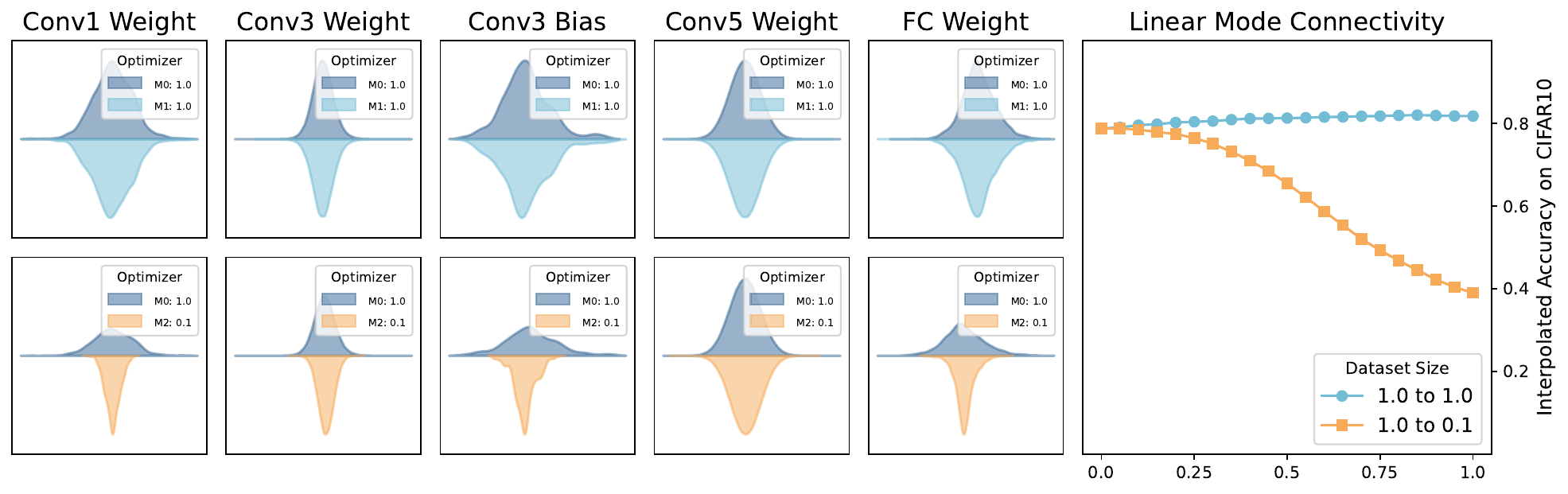}
    \caption{Different training conditions (data size) result in partially different weights and affect model interpolation.}
    \label{fig:factor-dsize}
\end{figure}

\begin{figure}[tb]
    \centering
    \includegraphics[width=\linewidth]{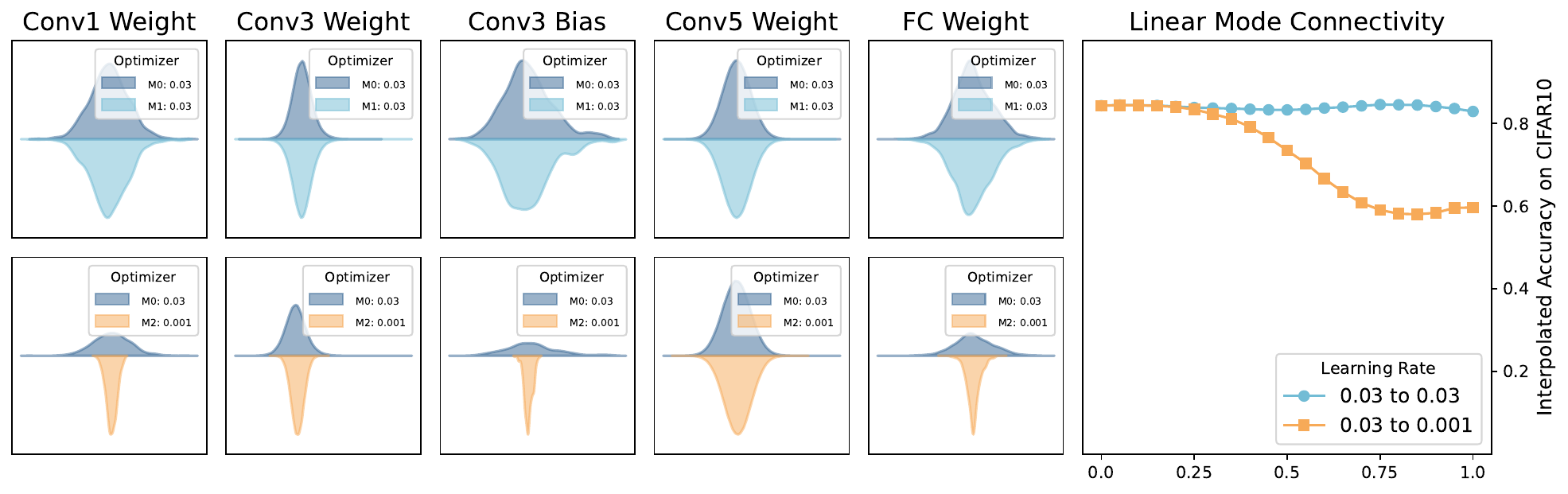}
    \caption{Different training conditions (learning rate) result in partially different weights and affect model interpolation.}
    \label{fig:factor-lr}
\end{figure}

Aside from the investigated factor, the other hyperparameters are set as same as~\Figref{fig:scope-mc-demo}, ensuring the training data remains original without any noise or sampling. Then, we could plot figures for each factor which are similar as \Figref{fig:scope-mc-demo}. \Figref{fig:factor-dsize} and \Figref{fig:factor-lr} illustrates different dataset size or learning rate  similarly lead to lead to weight scope mismatch and impact model interpolation.

Then we provide explanations for each of the columns in~\Tabref{tab:factor-barrier} .
\begin{enumerate}
    \item \textbf{KL D.}: this column computes the KL divergence of weight distributions between M0 and M2. More precisely, for each layer, we model the weight distribution as a Gaussian distribution and compute the KL divergence. The average result is then reported. Notably, the KL divergence between M0 and M1 is negligible, nearly zero, and therefore, we do not present the KL divergence between these models. This emphasizes that the divergence between models trained under identical conditions may approach zero.
    \item \textbf{Ba$(=)$}: this column calculates the barrier of the interpolation curve between model M0 and M1, and the definition of the barrier is provided in \Eqref{eq:loss_barrier} where we use the test error instead of the training loss.
    \item \textbf{Ba$(\neq)$}: this column calculates the barrier of the interpolation curve between M0 and M2, and the definition of the barrier is in \Eqref{eq:loss_barrier} where we use the test error instead of the training loss.
    \item \textbf{Diff}: it is the difference between ``Ba$(\neq)$'' and ``Ba$(=)$'', i.e., ``Ba$(\neq)$$-$Ba$(=)$''.
\end{enumerate}

\subsection{Different Weight Initialization}
In \Figref{fig:gaussian-dist},  we use two different popular initialization methods to train VGGBN16 models. The initialization method contains Kaiming uniform and Kaiming normal~\citep{he2015DelvingDeep}. Both of them show that the weight distribution follows a Gaussian distribution.

\subsection{Manual Scaling}

\begin{figure}[tb]
    \centering
    \includegraphics[width=\linewidth]{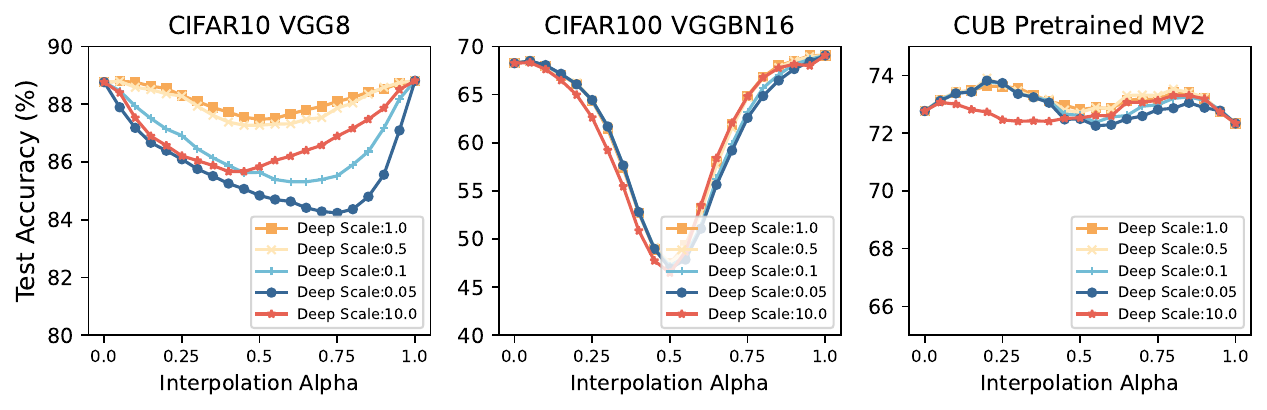}
    \caption{The linear interpolation curves between a standard model (Scale=1.0) and its scaled versions (Scale $\neq$ 1.0) are shown, in which the penultimate convolution layer is rescaled.}
    \label{fig:SCALE-50-appendix}
\end{figure}

\Figref{fig:manual-scale} illustrates the impact of mismatch in manually-designed weight scopes on linear interpolation. Our investigation covers pairs such as VGG8 on CIFAR10, VGGBN16 on CIFAR100, and Pretrained MobileNetV2 (MV2) on CUB. Specifically, we focus on the shallow layers, namely ``block0.0'' for VGG8 and VGGBN16, and ``backbone.0.1.conv.1'' for MV2, which represent the first convolution layers in their respective networks. Additionally, the study extends to deep layers: ``block3.0'' for VGG8, ``block4.6'' for VGGBN16, and ``backbone.0.17.conv.2'' for MV2, identified as the penultimate convolution layers. The consequent effects on linear interpolation are detailed in \Figref{fig:SCALE-50-appendix}. It is also observed that linear interpolation between two models with different weight distributions leads to deteriorated results.

\subsection{Mode Connectivity}
\begin{figure}[htb]
    \centering
    \includegraphics[width=\linewidth]{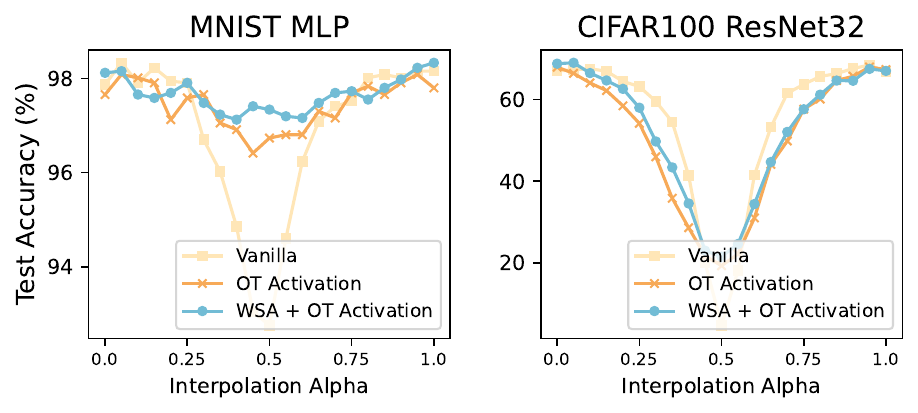}
    \caption{OTFusion on MNIST MLP and CIFAR100 ResNet-32.}
    \label{fig-suppl:ot-ws-other-dataset}
\end{figure}

In applying mode connectivity, our method is compared with model averaging (Vanilla), OTFusion~\citep{singh2020ModelFusion} and Git-rebasin~\citep{ainsworth2023GitRebasin} across different datasets and models. 
It is important to note that OTFusion and Git-rebasin are designed for different scenarios, necessitating separate comparative analyses. 
To make a comparison with OTFusion, we conduct experiments with VGG8 on SVHN and VGGBN8 on CIFAR10, as illustrated in~\Figref{fig:ot-ws}. 
Additionally, we present results for MLP on MNIST and ResNet-32 on CIFAR100 in \Figref{fig-suppl:ot-ws-other-dataset}. 
These results collectively demonstrate the benefits of weight scope alignment in enhancing mode connectivity within permutation-based model fusion methods.

Furthermore, we conduct comparisons with Git-rebasin~\citep{ainsworth2023GitRebasin} on the CIFAR-10 dataset using ResNet-18 and WRN. In this experiment, we train two models with different random seeds for 100 epochs and then compute their loss barriers on the test set using \Eqref{eq:loss_barrier}, with a learning rate of 0.01 and a weight decay of $10^{-4}$. The experimental results are displayed in \Figref{fig:fig_lmc_resnet_width_depth}.

We further assess the performance on RESISC45, DTD, GTSRB datasets included in VTAB~\citep{Zhai2019ALS}. Then, we compute the barrier of test accuracy as described in Section 5.2. A lower barrier indicates better interpolated performance. The results below demonstrate that WSA can reduce the barrier across the three datasets.

\begin{table}[]
    \centering
    \caption{The barrier of test accurary of ViTs finetuned on three datasets.}
    \label{tab:vtab_pretrained_model}
    \begin{tabular}{ccc}
    \toprule
    Dataset & Baseline & WSA \\ \midrule \midrule
    RESISC45 & 0.43\% & -0.22\% \\
    DTD & -0.50\% & -0.93\% \\
    GTSRB & 0.90\% & 0.79\% \\
    \bottomrule
    \end{tabular}
\end{table}

\subsection{Federated Learning}
\begin{figure}[hb]
    \centering
    \includegraphics[width=\linewidth]{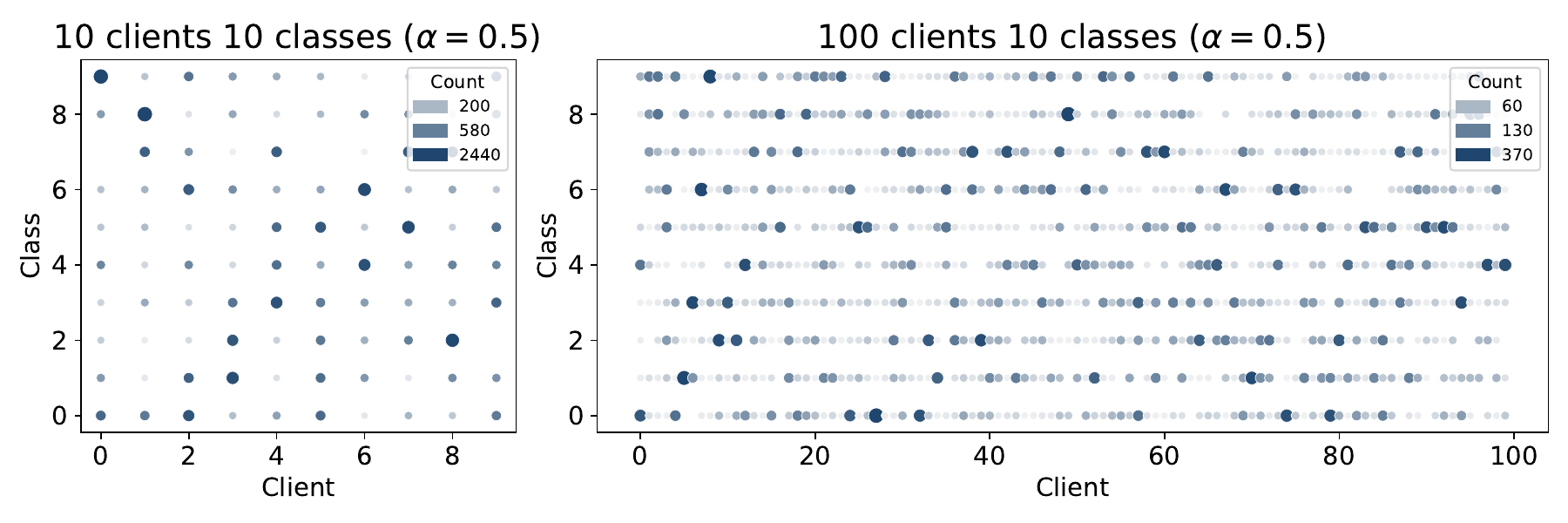}
    \caption{Client distribution of 10 clients and 100 clients smapled from dirichelt distribution ($\alpha=0.5$).}
    \label{fig:data_dist_0_5}
\end{figure}

\begin{figure}[hb]
    \centering
    \includegraphics[width=\linewidth]{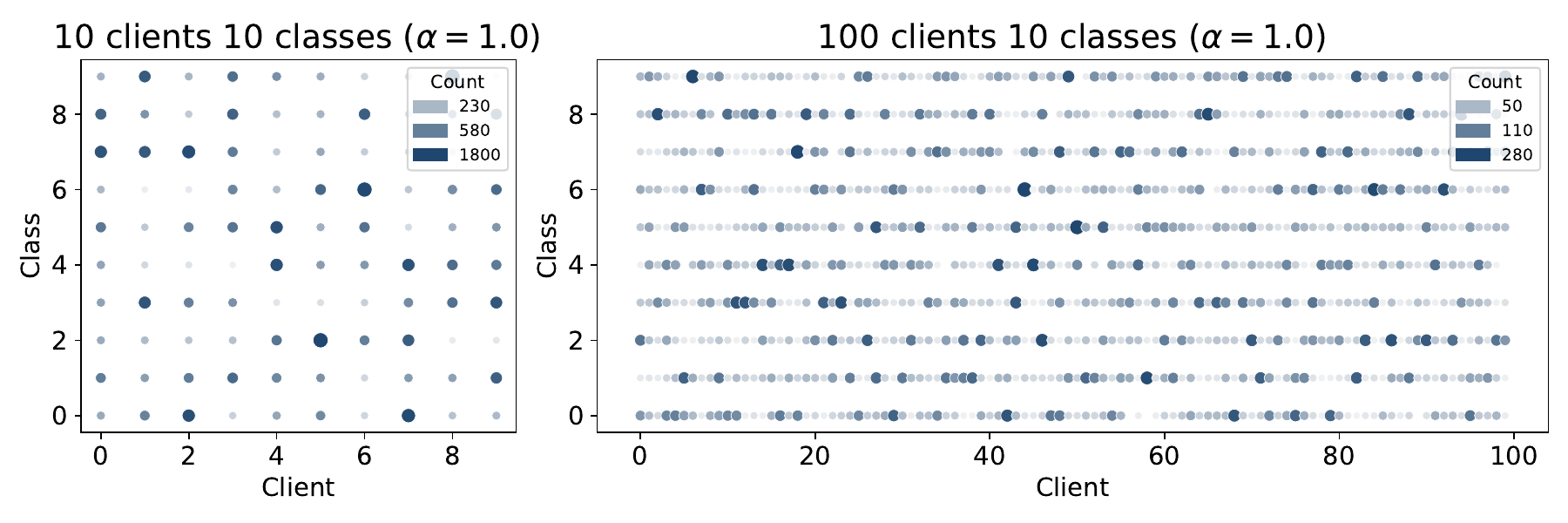}
    \caption{Client distribution of 10 clients and 100 clients smapled from dirichelt distribution ($\alpha=1.0$).}
    \label{fig:data_dist_1}
\end{figure}

\begin{figure*}[htb]
    \centering
    \includegraphics[width=\linewidth]{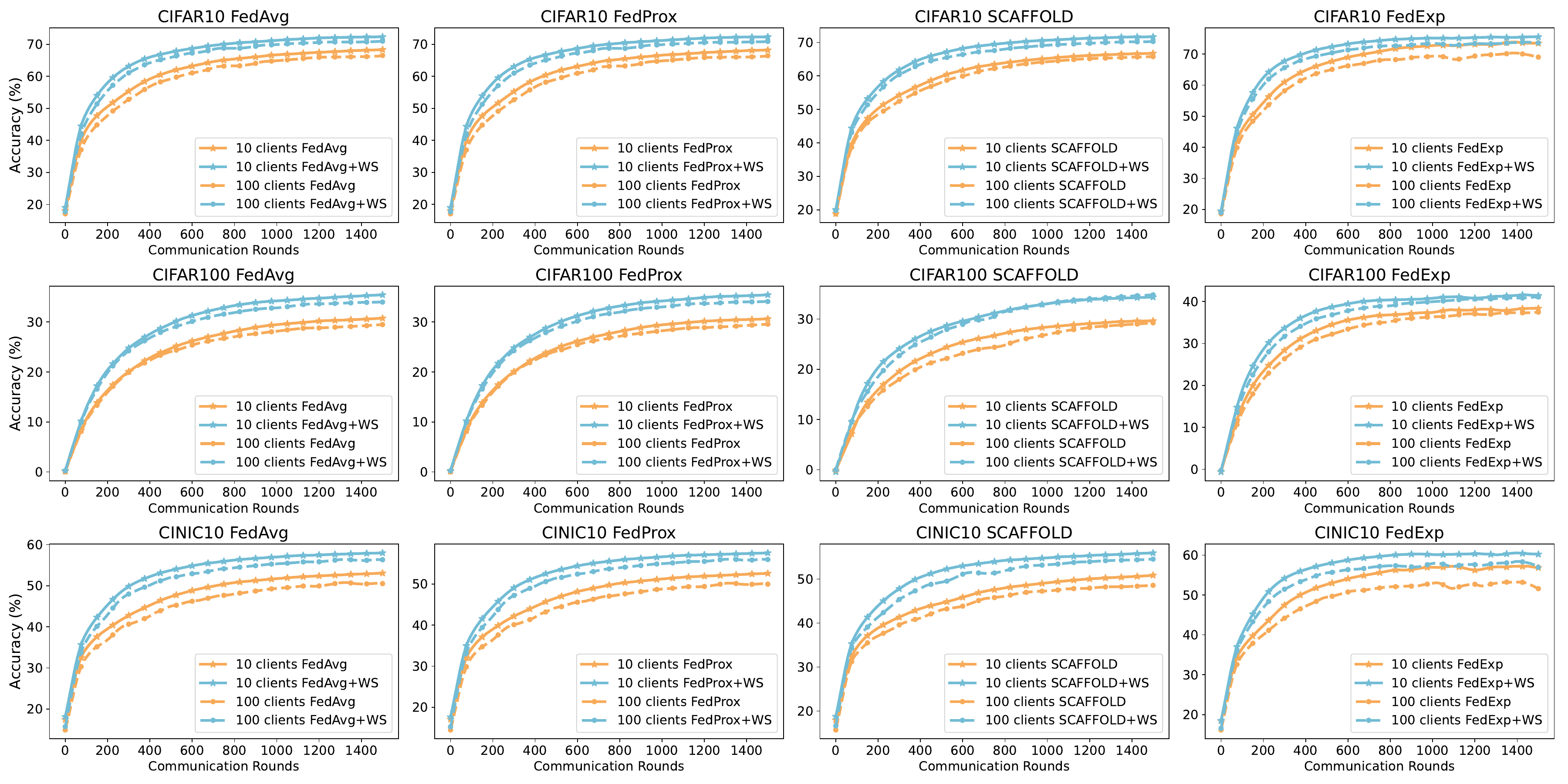}
    \caption{Convergence curve of test accurarcy across various settings in FL.}
    \label{fig:convergence_baseline}
\end{figure*}

To simulate various clients, we employ a Dirichlet distribution to allocate data across clients for the CIFAR-10, CIFAR-100, and CINIC-10 datasets~\citep{hsu2019MeasuringEffects}, utilizing the parameter $\alpha$ to control data heterogeneity. \Figref{fig:data_dist_0_5} and~\Figref{fig:data_dist_1} depict the data distribution among clients for $\alpha$ values of 0.5 and 1.0, respectively, with lower $\alpha$ values indicating increased data heterogeneity.
The training convergence curves in 10 clients and 100 clients setting are illustrated in~\Figref{fig:convergence_baseline}.

Our experiments are conducted using the FedExP framework~\footnote{\url{https://github.com/Divyansh03/FedExP}}. 
The experimental setup includes a learning rate of 0.01, weight decay of $10^{-4}$, a decay of the learning rate by 0.998 in each round, a maximum gradient norm of 10, a fusion alpha of 1.0, and 20 local steps for all models and datasets. Specifically for FedProx, $\mu$ is set to 0.1, 1, and 0.001 for CIFAR-10, CIFAR-100, and CINIC-10, respectively. In the case of FedExp, $\epsilon$ is maintained at 0.001 which is identified as the optimal value for all three datasets in~\citep{jhunjhunwala2023FedExPSpeeding}.

\begin{table}[]
    \centering
    \caption{Hyperparameter sensitivity in FL.}
    \label{tab:hyperparameter}
    \begin{tabular}{ccc}
    \toprule
    $\lambda$ & Test Acc (Epoch 100) & Test Acc (Epoch 1000) \\ \midrule \midrule
    0 & 42.34 & 63.06 \\
    1 & 47.59 & 67.52 \\
    5 & 51.99 & 67.72 \\
    10 & 53.18 & 66.58 \\
    50 & 52.90 & 62.68 \\
    \bottomrule
    \end{tabular}
\end{table}

\subsubsection{Additional Experiments}

\textbf{Hyperparameter sensitivity} In \Eqref{eq:loss_fedavg_ws}, $\lambda$ is the hyperparameter that controls the strength of weight scope regularization. We explore the sensitivity of the hyperparameter $\lambda$ by testing values in [1, 5, 10, 50] within the context of federated learning. The results below indicate that when $\lambda=5$, the performance is optimal. However, setting $\lambda$ significantly higher, initially leads a rapid performance increase as shown the performance at the 100-th epoch. However, the final convergence performance is actually lower. Intuitively, applying stronger weight scope regularization may cause the model to lose flexibility, akin to large weight decay.

\end{document}